\documentclass[a4paper, 10pt, twocolumn, twoside]{article}

\usepackage{ISARC}

\usepackage{lscape}
\usepackage{hologo}

\begin{document}

\linespread{0.5}

\title{One-class Damage Detector Using Deeper Fully Convolutional Data Descriptions for Civil Application}

\author{Takato Yasuno$^{1}$, Masahiro Okano$^1$ and Junichiro Fujii$^1$}

\affiliation{
$^1$Research Institute for Infrastructure Paradigm Shift
}

\email{
\href{mailto:e.author1@aa.bb.edu}{tk-yasuno@yachiyo-eng.co.jp}, 
\href{mailto:e.author1@aa.bb.edu}{ms-okano@yachiyo-eng.co.jp}, 
\href{mailto:e.author1@aa.bb.edu}{jn-fujii@yachiyo-eng.co.jp}
}

\maketitle 
\thispagestyle{fancy} 
\pagestyle{fancy}

\begin{abstract}
Infrastructure managers must maintain high standards to ensure user satisfaction during the lifecycle of infrastructures. Surveillance cameras and visual inspections have enabled progress in automating the detection of anomalous features and assessing the occurrence of deterioration. However, collecting damage data is typically time consuming and requires repeated inspections. The one-class damage detection approach has an advantage in that normal images can be used to optimize model parameters. Additionally, visual evaluation of heatmaps enables us to understand localized anomalous features. 
The authors highlight damage vision applications utilized in the robust property and localized damage explainability.
First, we propose a civil-purpose application for automating one-class damage detection reproducing a fully convolutional data description (FCDD) as a baseline model. 
We have obtained accurate and explainable results demonstrating experimental studies on concrete damage and steel corrosion in civil engineering.  
Additionally, to develop a more robust application, we applied our method to another outdoor domain that contains complex and noisy backgrounds using natural disaster datasets collected using various devices. Furthermore, we propose a valuable solution of 
deeper FCDDs focusing on other powerful backbones to improve the performance of damage detection and implement ablation studies on disaster datasets. 
The key results indicate that the deeper FCDDs outperformed the baseline FCDD on datasets representing natural disaster damage caused by hurricanes, typhoons, earthquakes, and four-event disasters.
\end{abstract}


\section{Introduction}
\label{sec:Introduction}

\subsection{Related Works on Vision-based Anomaly Detection}
Over the past decade, anomaly detection techniques have attracted significant attention to widespread domain of applications assisted by the methodologies of machine learning and deep learning. Previous survey papers provided fruitful systematic overviews \cite{Chandola2009}\cite{Chalapathy2019}\cite{Ruff2020}\cite{Yuan2022}, focused on the model property, application domain, and trustworthiness to be more interpretable, fair, robust, and privacy settings. Specifically, vision-based deep learning applications have emerged by two driving forces: computing accessibility and digitalized society that accelerate the creation of many datasets annotated with several class labels. There has been over 20  datasets of surface damage for industrial products that have focused on various materials: steel, metal, aluminum, tile, fabric, printed board, solar panel, and civil infrastructures: concrete, road, pavement, bridge, and rail \cite{Saberironaghi2023}. The construction domain is no exception, image-based structural health monitoring and visual inspection techniques have been facilitated using deep learning algorithms \cite{Payawal2023}\cite{Wang2019}. Visual structural datasets enable to promote the development of widespread applications, over 80 studies towards the infrastructure damage: deterioration, displacement, and exfoliation \cite{Bianchi2022}.
This paper highlights the damage vision application utilized in the robust property and localized damage explainability.

\begin{figure*}[h]
\centering
\includegraphics[width=0.85\textwidth]{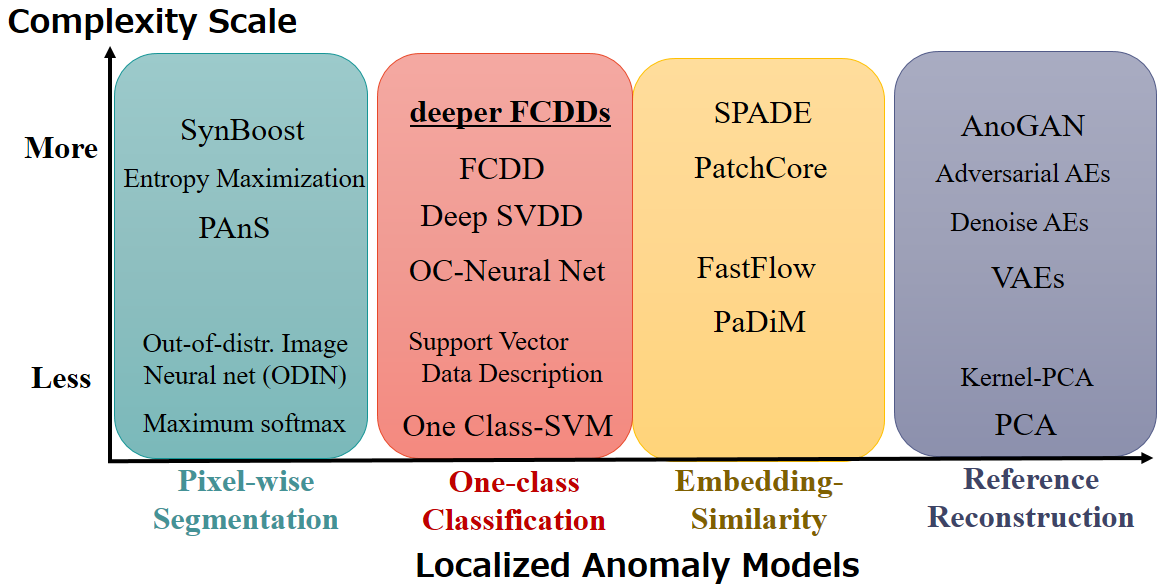}
\caption{\label{fig:models}Our proposed deeper FCDDs via the existing anomaly detection models.}
\end{figure*}
\begin{figure*}[h]
\centering
\includegraphics[width=0.8\textwidth]{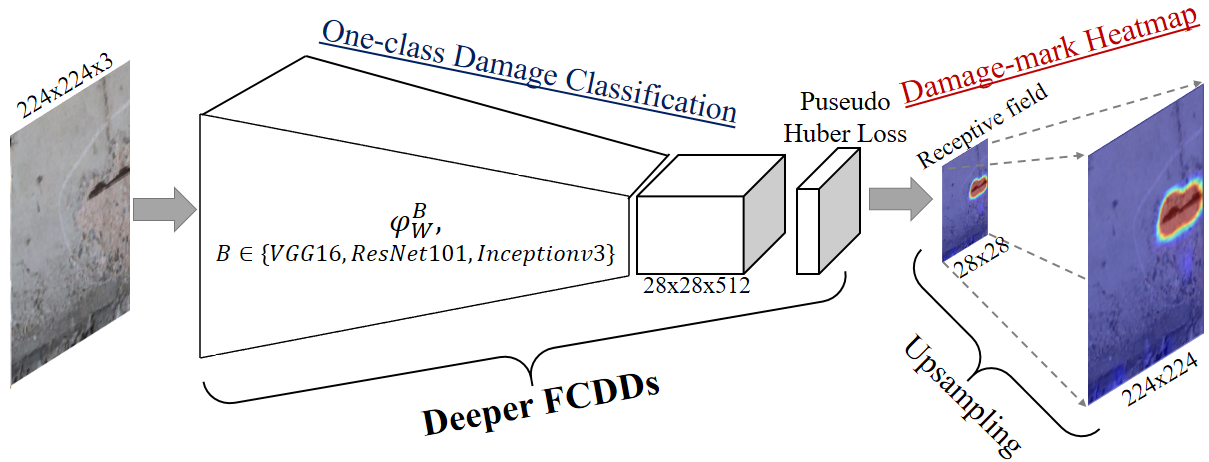}
\caption{\label{fig:overview}Overview of civil damage classification using deeper FCDDs and a damage heatmap.}
\end{figure*}

As shown in Figure \ref{fig:models}, modern anomaly detection approaches can be divided into four categories: pixel-wise segmentation, one-class classification, patch-wise embedding-similarity, and reconstruction-based models. Inspired with \cite{Ruff2020}, these anomaly detection approaches are reviewed in a unified manner progressing from less to more complexity scale through several categories of localization models. Firstly, anomaly detection approaches based on less complexity scale include the one-class support vector machine (OC-SVM) \cite{Chalapathy2018}, support vector data description (SVDD) \cite{Tax2014}, principal component analysis (PCA) \cite{Hawkins1974}, and kernel-PCA \cite{Hoffmann2007}. Anomaly detection approaches based on more complexity scale include deep SVDD \cite{Ruff2018}, fully convolutional data description (FCDD) \cite{Liznerski2021}, variational autoencoder (VAE) \cite{Kingma2019}\cite{An2015}, and adversarial auto-encoders (AAE) \cite{Zhou2017}. However, reconstruction-based models cannot always reconstruct synthetic outputs well based on susceptibility to background noise. In contrast, one-class classification models depend on neither synthetic reconstruction nor probabilistic assumptions; therefore, they may be more robust anomaly detectors. 

Pixel-wise segmentation approach is detecting unknown objects in semantic segmentation for perception in the automated driving \cite{Chan2022}. Anomaly segmentation methods contains the maximum softmax probability \cite{Hendrycks2017}, out-of-distribution image  detector in Neural networks (ODIN) \cite{Liang2018}, SynBoost \cite{Biase2021}, entropy maximization \cite{Chan2021}, and PAnS \cite{Fontanel2021}. However, semantic segmentation models used to require lots of annotation cost and heavy memory for training and prediction. This scope of pixel-wise localization must be over-specification for the aim of light applicability in thousands of outdoor scenes.   
Separately, patch-wise embedding approach enables to minimize the background noise per each patch image. To localize the anomalous feature in a patch image, patch-wise embedding-similarity models perform that the normal reference can be the sphere feature containing embeddings from normal images. In this case, anomaly score is the distance between embedding vectors of a test image and reference vectors representing normality from the dataset. Embedding-similarity based models includes the SPADE \cite{Cohen2020}, PaDiM \cite{Thomas2020}, PatchCore \cite{Roth2021}, FastFlow \cite{Yu2021}. However, these models are based on supervised learning that additionally requires optimization algorithms such as a greedy coreset selection, a nearest neighbor search on a set of normal embedding vectors, so the inference complexity scales linearly to the size of training dataset. In contrast, one-class classification approach can learn efficiently using rare class of imbalanced dataset with fewer scale for damage detection in civil applications.  

\subsection{Civil Application and Robustness during Natural Disasters}
In civil applications, we have performed the anomaly detection task by focusing on various types of infrastructure damage, including damage on bridge slabs using human eye inspection, dam embankments using auto-flight drone images, and fallen objects on roads using an internet protocol (IP) camera.   
For example, we proposed a bridge slab anomaly detector using a U-Net generator with a patch discriminator containing AAEs and an OC-SVM \cite{YasunoIWSHM2021}. 
Additionally, we proposed a concrete damage detection method using an auto-flight UAV based on cycleGAN and morphology analysis for computing anomaly scores \cite{YasunoISARC2020}.
We also proposed a pipeline combining VAE reconstruction with an isolation forest for detecting fallen objects on road surfaces after a preprocessing translation operation using pix2pix \cite{YasunoJSAI2022}. 
However, we could not completely reconstruct the synthetic surface images of concrete and asphalt outdoors. This is because sunshine and shadow conditions are not always consistent and unavoidable noise, such as green moss under wet conditions and dirty surfaces following decades of public service, frequently exists. 
Limited data collected under unified conditions cannot facilitate stable training for reconstruction approaches considering the wide variety of background noise introduced by seasonal changes in addition to different specifications of outdoor infrastructures.

Regarding hazard recognition during and following natural disasters, we have performed the anomaly detection task by focusing on disaster damage, such as fallen trees following typhoons and broken building roofs, using aerial photographs and winter snowy road monitoring using IP camera surveillance.
We proposed a pipeline for normal land use and typhoon damage classification and an intensity-scaled heatmap based on a composite matrix of class probabilities per patch \cite{asahi2020}.  
Additionally, to address the winter road safety problem under snowy conditions and make decisions regarding early morning snowplow application, we proposed a pipeline that performs road surface translation without mobility using pix2pix and semantic segmentation of snow hazard indices on road regions without background snow \cite{YasunoAAI2021}.
Furthermore, we proposed adding a preprocessing operation of night-to-day translation from a lit road at night to road surface conditions during the day to compute a snow coverage index based on night vision \cite{YasunoJSAI2021}.   
However, these pipelines were combined with a deep reconstruction algorithm for a synthetic normal surface and shallow machine learning algorithm for computing anomaly scores.
These combined pipelines could not consistently achieve high performance based on the limitations of simultaneous accuracy control.      
For more robust applications, an end-to-end solution for detecting anomalies based on a convolutional damage data description and damage heatmap visualization is required. 

In this paper, we propose a civil-purpose application to automate one-class damage detection using an FCDD. We also visualize damage features using the Gaussian upsampling of the receptive field of a fully convolutional network (FCN). Figure \ref{fig:overview} provides an overview of infrastructure damage classification using an FCDD and upsampling-based heatmap explanation. 
Additionally, to develop a more robust application, we applied our method to an outdoor domain containing complex and noisy backgrounds such as natural disaster damage owing to hurricanes, typhoons, earthquakes, and four-event disasters. These disaster images were collected using various modes, including satellite imagery, aerial photography, drone-based systems, and panoramic 360 cameras.   
Furthermore, to improve the performance of damage detection, we propose deeper FCDDs incorporating other deeper backbones such as VGG16, ResNet101, and Inceptionv3. We conducted ablation studies and compared the results to those of the initial baseline FCN.

\section{Damage Detection Method}

\subsection{One-class Damage Classification Using Deeper FCDDs}
Let $X_i$ be the $i$-th image with a size of $h\times w$, and let $c$ be the center of the hypersphere boundary between the inlier normal region and outlier anomalous region. We consider the number of training images, as well as the weight $W$ of the FCN. The deep SVDD objective function \cite{Ruff2018} is formulated as a minimization problem for deep support vector data description as follows:
\begin{equation}
\min_{W} \frac{1}{n} \sum_{i=1}^{n} \| \varphi^B_W(X_i) - c \|^2,
\end{equation}
where denotes the $\varphi^B_W(X_i)$ mapping of the deeper CNN to the backbone $B$ based on the input image. The one-class classification model is formulated as follows using the cross-entropy loss function:
\begin{equation}
\begin{split}
\mathcal{L}_{DeepSVDD} = &- \frac{1}{n} \sum_{i=1}^{n} (1-y_i) \log \ell (\varphi^B_W(X_i))\\
                    &+ y_i \log [ 1 -  \ell (\varphi^B_W(X_i)) ],
\end{split}
\end{equation}
where $y_i =1$ denotes the anomalous label of the $i$-th image and $y_i =0$ denotes the normal label of the $i$-th image. For a more robust loss formulation, the pseudo-Huber loss function was introduced \cite{Ruff2021icml} in Equation (2). We let $\ell(z)$ be the loss function and define the pseudo-Huber loss as follows:   
\begin{equation}
\ell(z) = \exp(-H(z)),~ H(z) = \sqrt{\|z\|^2 + 1} -1.
\end{equation}
By substituting Equation (2) into Equation (3), we obtain the following expression:
\begin{equation}
\begin{split}
(2) \equiv & - \frac{1}{n} \sum_{i=1}^{n} (1-y_i) H (\varphi^B_W(X_i)) \\
                   & + y_i \log [ 1 -  \exp\{ -H(\varphi^B_W(X_i)) \} ].
\end{split}
\end{equation}
Therefore, the deeper FCDD loss function can be formulated as follows:
\begin{equation}
\begin{split}
& \mathcal{L}_{deeperFCDD} = \frac{1}{n} \sum_{i=1}^{n} \frac{(1-y_i)}{uv} \sum_{x,y} H_{x,y} (\varphi^B_W(X_i)) \\ 
                           & - y_i \log \left[ 1 -  \exp\left\{ \frac{-1}{uv} \sum_{x,y} H_{x,y} (\varphi^B_W(X_i)) \right\} \right],
\end{split}
\end{equation}
where $H_{x,y}(z)$ are the elements $(x,y)$ of the receptive field with a size of $u\times v$ under the deeper FCDD. 
The anomaly score $S_i$ of the $i$-th image is expressed as the sum of all elements of the receptive field as follows:
\begin{equation}
S^B_i = \sum_{x,y} H_{x,y} (\varphi^B_W(X_i)),~i=1,\cdots,n.
\end{equation}
In this study, we constructed a baseline FCDD with an initial backbone $B=0$ and performed CNN27 mapping $\varphi^0_W(X_i)$ from input images $X_i$ in civil datasets. We also present deeper FCDDs focusing on the elaborate backbones $B\in \{$VGG16, ResNet101, Inceptionv3$\}$ with a mapping operation $\varphi^{B}_W(X_i)$ to achieve a more robust detection. We also present several ablation studies on disaster datasets. 

\subsection{Damage Mark Heatmap Upsampling from the Receptive Field}
CNN models with millions of shared parameters have achieved satisfactory performance for anomaly detection. However, the reasons for their impressive performance remain unclear. Heatmap visualization techniques can largely be divided into masked sampling and activation map approaches. The former category includes occlusion sensitivity \cite{Zeiler2013} and local interpretable model-agnostic explanations \cite{Ribeiro2016}. The main merit of this approach is that it does not require in-depth knowledge of network architecture, but its main disadvantage is that it requires iterative computations per image and additional running time for local partitioning, masked sampling, and output prediction. The last category includes activation maps such as class activation maps (CAMs) \cite{Zhou2015} and gradient-based extension (Grad-CAM) \cite{Selvaraju2017}. Weighting the feature maps of CAMs is ineffective because it limits global average pooling and full connection effectiveness in the final layer of a CNN. The main advantage of the gradient approach is that it can be applied to any layer of a CNN; therefore, it has significantly improved applicability. However, the main disadvantage is that parallel computation resources and a moderate running time are required for generating a gradient-based heatmap. 

For civil-purpose applications, we selected the receptive field upsampling approach \cite{Liznerski2021} to visualize anomalous damage features using an upsampling-based activation map with Gaussian upsampling from the receptive field of the FCN. The main advantages of the upsampling approach include reduced computational resource requirements and lower running times. The proposed upsampling algorithm generates a full-resolution anomaly heatmap from the input of a low-resolution receptive field $u\times v$. 
Let $H\in \mathbb{R}^{u\times v}$ be a low-resolution receptive field (input), and let $H'\in \mathbb{R}^{h\times w}$ be a full-resolution damage heatmap (output).
We define a 2D Gaussian distribution $G_2(m_1,m_2,\sigma)$ as follows: 
\begin{equation}
\begin{split}
& [G_2(m_1,m_2,\sigma)]_{x,y} \\ 
& \equiv \frac{1}{2\pi\sigma^2}\exp\left(-\frac{(x-m_1)^2+(y-m_2)^2}{2\sigma^2}\right).  
\end{split}
\end{equation}
The Gaussian upsampling algorithm from the receptive field is then implemented as follows:
\begin{enumerate}
\item $H' \leftarrow 0 \in \mathbb{R}^{h\times w}$
\item for all output pixels $d$ in $H \leftarrow 0 \in \mathbb{R}^{u\times v}$
\item  $u(d) \leftarrow$ is upsampled from a receptive field of $d$
\item  $(c_1(u),c_2(u)) \leftarrow$ is the center of the field $u(d)$
\item  $H' \leftarrow H' + d\cdot G_2(c_1,c_2,\sigma)$
\item end for
\item return $H'$ 
\end{enumerate}
Based on the experiments on various datasets, we set the size of the receptive field to $28\times 28$ as a practical value. To generate a damage heatmap, we must unify the display range corresponding to the anomaly scores ranging from the minimum to maximum value. To strengthen the damage regions and highlight the damage marks, we define a display range of [min. max./4], whose quartile parameter is 0.25. Therefore, the histogram of anomaly scores has a long-tailed shape. If we include the complete anomaly score range, then the color would be weakened to blue or yellow on the maximum side.

\begin{table}[h]
\caption{\label{tab:datacivil}Damage datasets used for the inspection of roads, bridges, and dams.}
\centering
\begin{tabular}{c|c|r|r}
Dataset & Size & Normal & Anomalous \\\hline
Pavement crack & $256^2$ & 3,500 &1,826 \\
Bridge rebar exposure & $224^2$ & 306 & 230 \\
Bridge steel corrosion & $64^2$ & 2,400 & 5789 \\
Dam exfoliation, janka & $256^2$ & 1,075 & 247 \\
\end{tabular}
\end{table}
\begin{table*}[h]
\caption{\label{tab:cnn27}Layer types and shapes in the baseline FCDD architecture on CNN27.}
\centering
\begin{tabular}{c|c|c|c|c}
No. & Layer type & Output shape $(S,S,C)$ & Kernel & Learnable parameters.  \\\hline
1 & Input & 224,224,3 & -- & -- \\
2-4 &	Conv1-BN-Relu1 & 224,224,64 & 3 & 1,792 \\
5 & Maxpool1 & 112,112,64 & -- & --\\
6-8 &	Conv2-BN-Relu2 & 112,112,128 & 3 & 73,856 \\
9 & Maxpool2 & 56,56,128 & -- & -- \\
10-12 & Conv3-BN-Relu3 & 56,56,256 & 3 & 295,168 \\
13-15 & Conv4-BN-Relu4 & 56,56,256 & 3 & 295,168 \\
16 & Maxpool3 & 28,28,256 & -- & -- \\
17-19 & Conv5-BN-Relu5 & 28,28,512 & 3 & 1,180,160 \\
20-22 & Conv6-BN-Relu6 & 28,28,512 & 3 & 1,180,160 \\
23-25 & Conv7-BN-Relu7 & 28,28,512 & 3 & 1,180,160 \\
26 & Conv8 & 28,28,512 & 1 & 264,192 \\
27 & Pseudo Huber loss & -- & -- & -- \\ \hline			
-- & total & Learnables & -- & 4.4M \\
\end{tabular}
\end{table*}
\begin{table*}[h]
\caption{\label{tab:accCivil}Accuracy of damage detection using the baseline FCDD for roads, bridges, and dams.}
\centering
\begin{tabular}{c|c|c|c|c|c}
Model   & Dataset & AUC & $F_1$ & Precision & Recall \\\hline
           &SDNET Pavement crack & \textbf{0.8955} & 0.7104 & 0.6209 & \textbf{0.8301} \\
\textbf{baseline}& Bridge rebar exposure & \textbf{0.9649} & 0.9052 & 0.8775 & \textbf{0.9347} \\
\textbf{FCDD} & Bridge steel corrosion & \textbf{0.9889} & 0.8803 & 0.7972 & \textbf{0.9827} \\
           & Dam exfoliation, janka & \textbf{0.9249} & 0.7831 & 0.7469 & \textbf{0.8231} \\
\end{tabular}
\end{table*}


\section{Application Results using the Baseline FCDD}

\subsection{Damage Datasets for Civil Engineering}
As summarized in Table~\ref{tab:datacivil}, we demonstrate a civil-purpose application through experimental studies on pavement cracks from the SDNET dataset \cite{Dorafshan2018}, bridge rebar exposure and steel paint peeling, volt nut corrosion, and dam embankment janka.

As summarized in Table~\ref{tab:cnn27}, we constructed an FCN as the initial backbone with 27 layers and 4.4 million learnable parameters, which was termed as CNN27 and contained either a Conv-BN-ReLU or Maxpool activation function. This initial FCN used for the prototype detector had neither a skip layer nor residual layer. 
Table~\ref{tab:accCivil} summarizes the accuracy values of one-class damage detection when applied to the damage dataset for roads, bridges, and dams. The area under the curve (AUC) and recall values are considerably high, suggesting that the FCDD is suitable for civil damage inspection applications.

\begin{figure*}[h]
\centering
\includegraphics[width=0.28\textwidth]{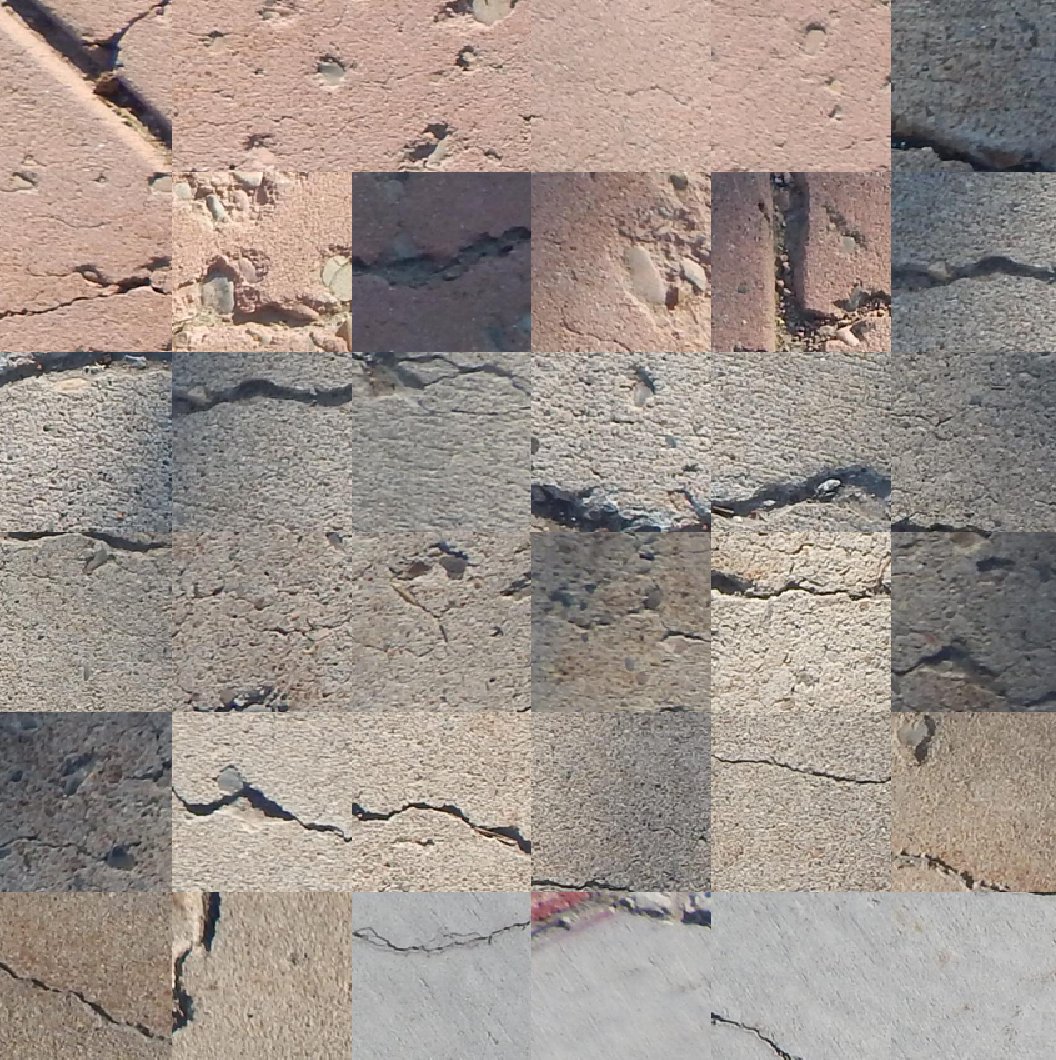}
~\includegraphics[width=0.28\textwidth]{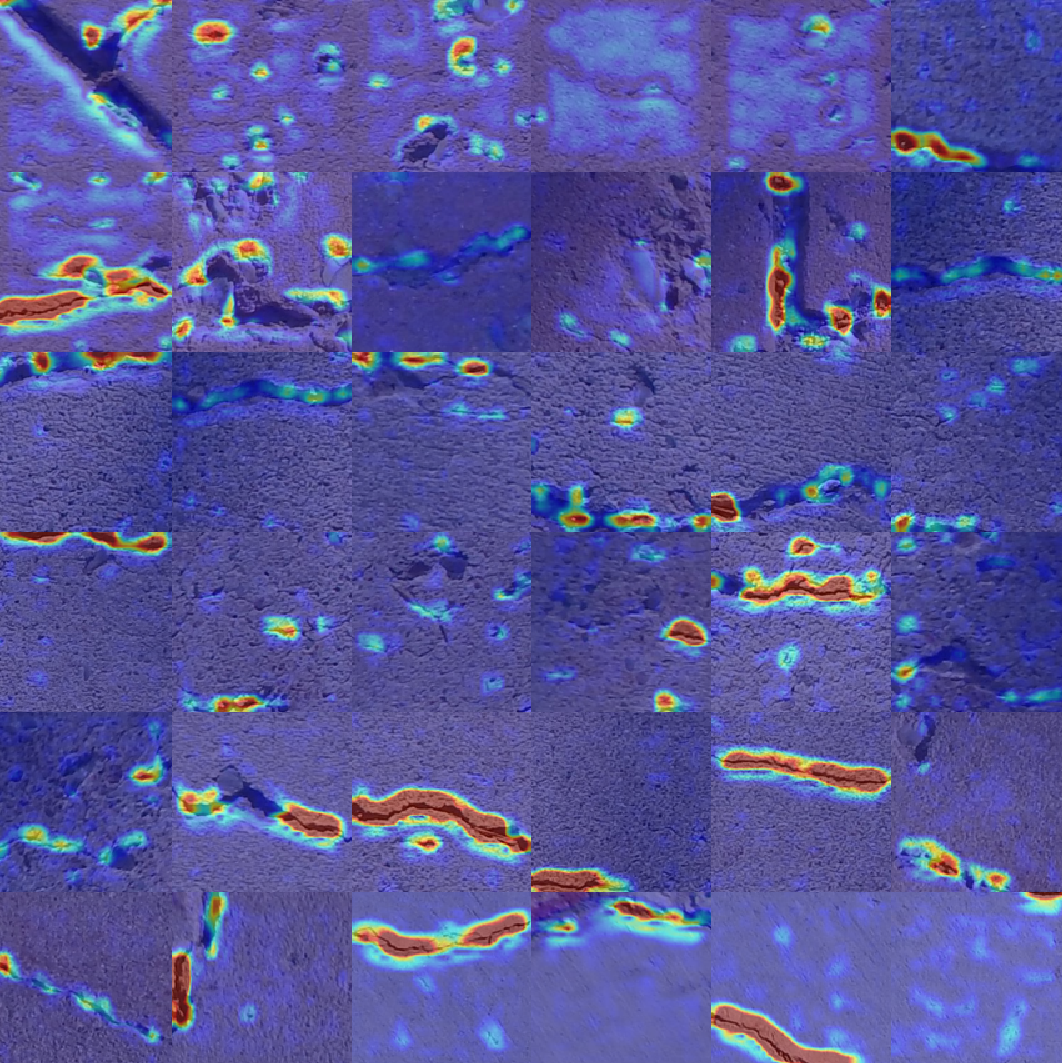}
~\includegraphics[width=0.35\textwidth]{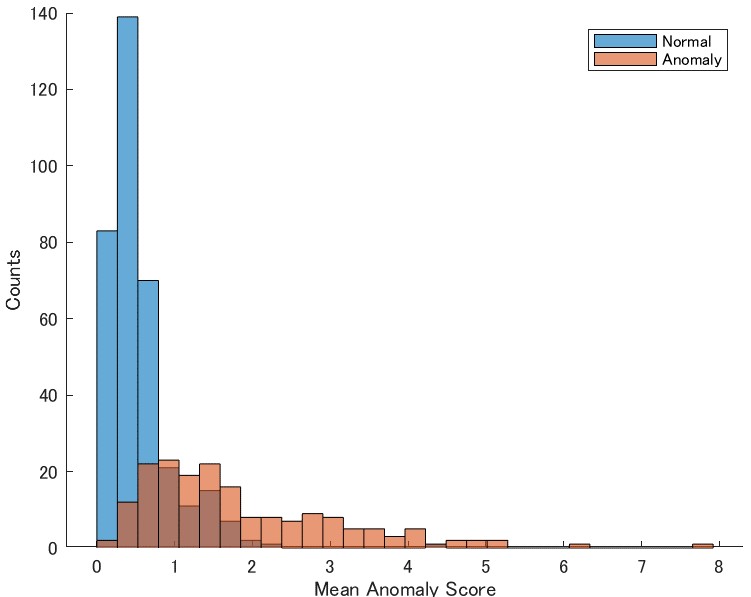}
\caption{\label{fig:rawheatPave}Input images (left) of SDNET pavement cracks, results for damage mark heatmaps (middle), and a histogram (right) corresponding to the baseline FCDD based on CNN27.}
\end{figure*}
\begin{figure*}[h]
\centering
\includegraphics[width=0.28\textwidth]{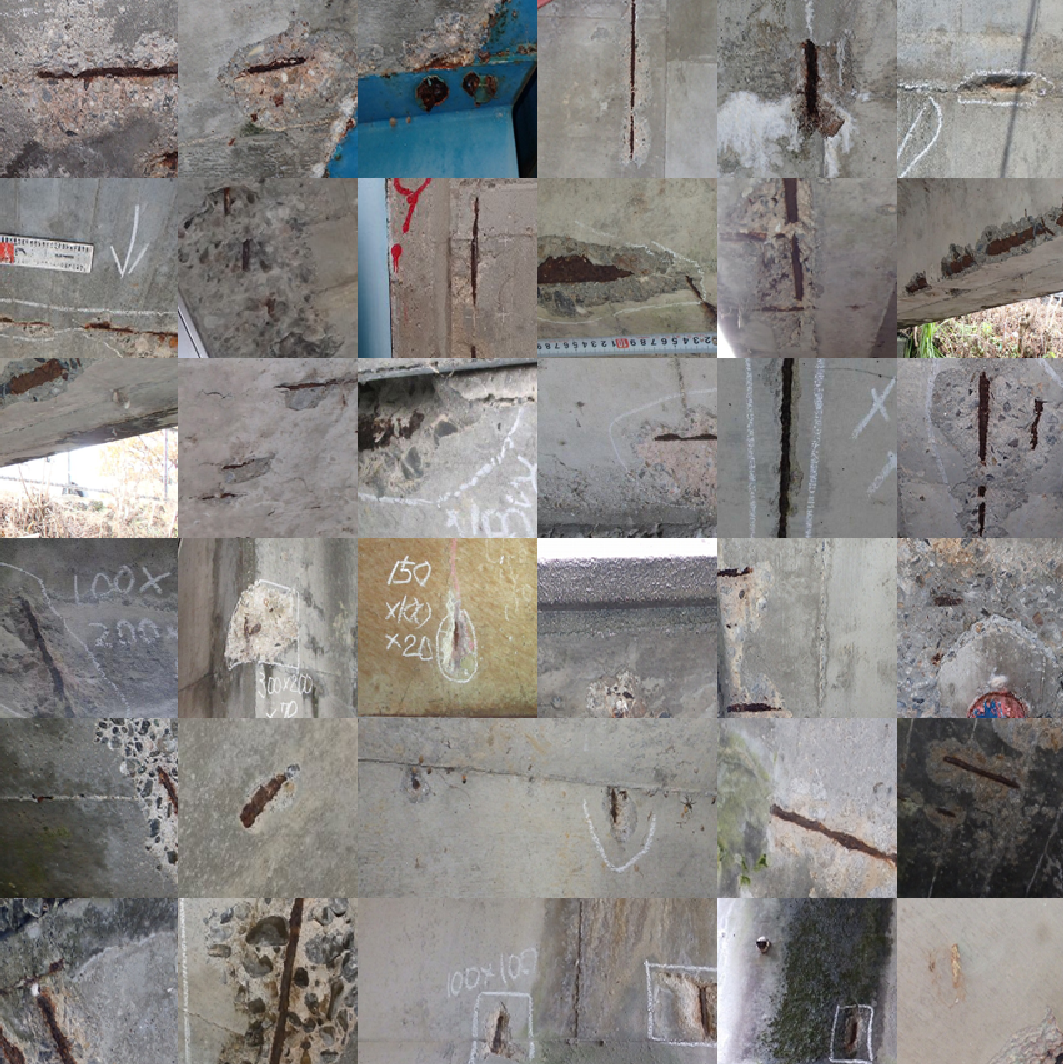}
~\includegraphics[width=0.28\textwidth]{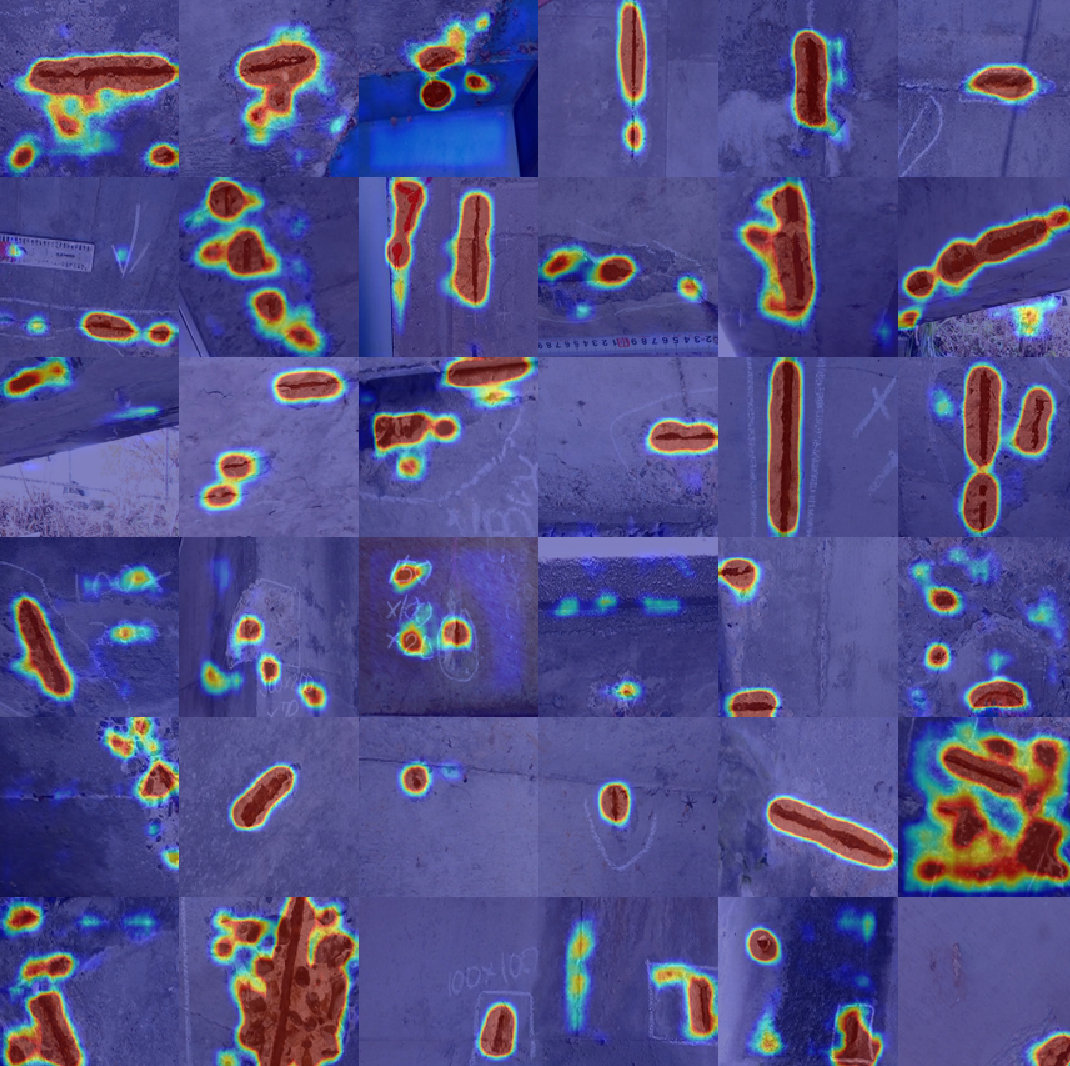}
~\includegraphics[width=0.35\textwidth]{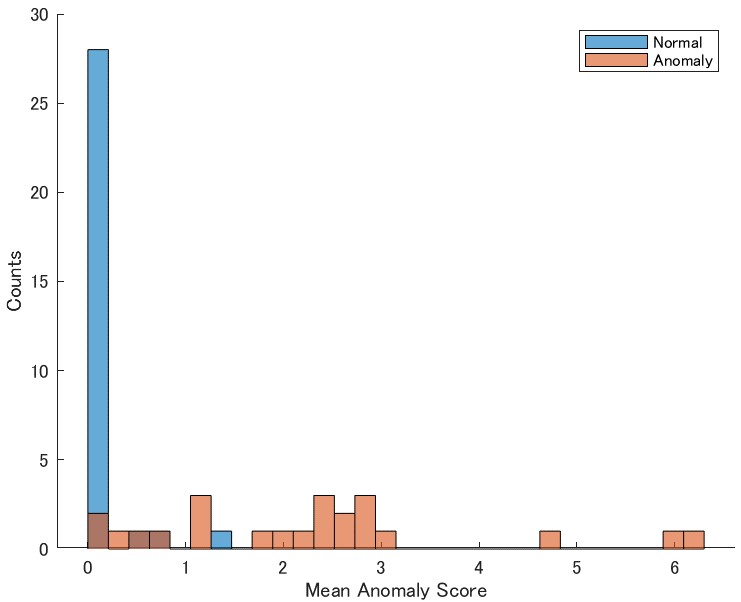}
\caption{\label{fig:rawheatRebar}Input images (left) of bridge rebar exposure, results for damage mark heatmaps (middle), and a histogram (right) corresponding to the baseline FCDD based on CNN27.}
\end{figure*}
\begin{figure*}[h]
\centering
\includegraphics[width=0.28\textwidth]{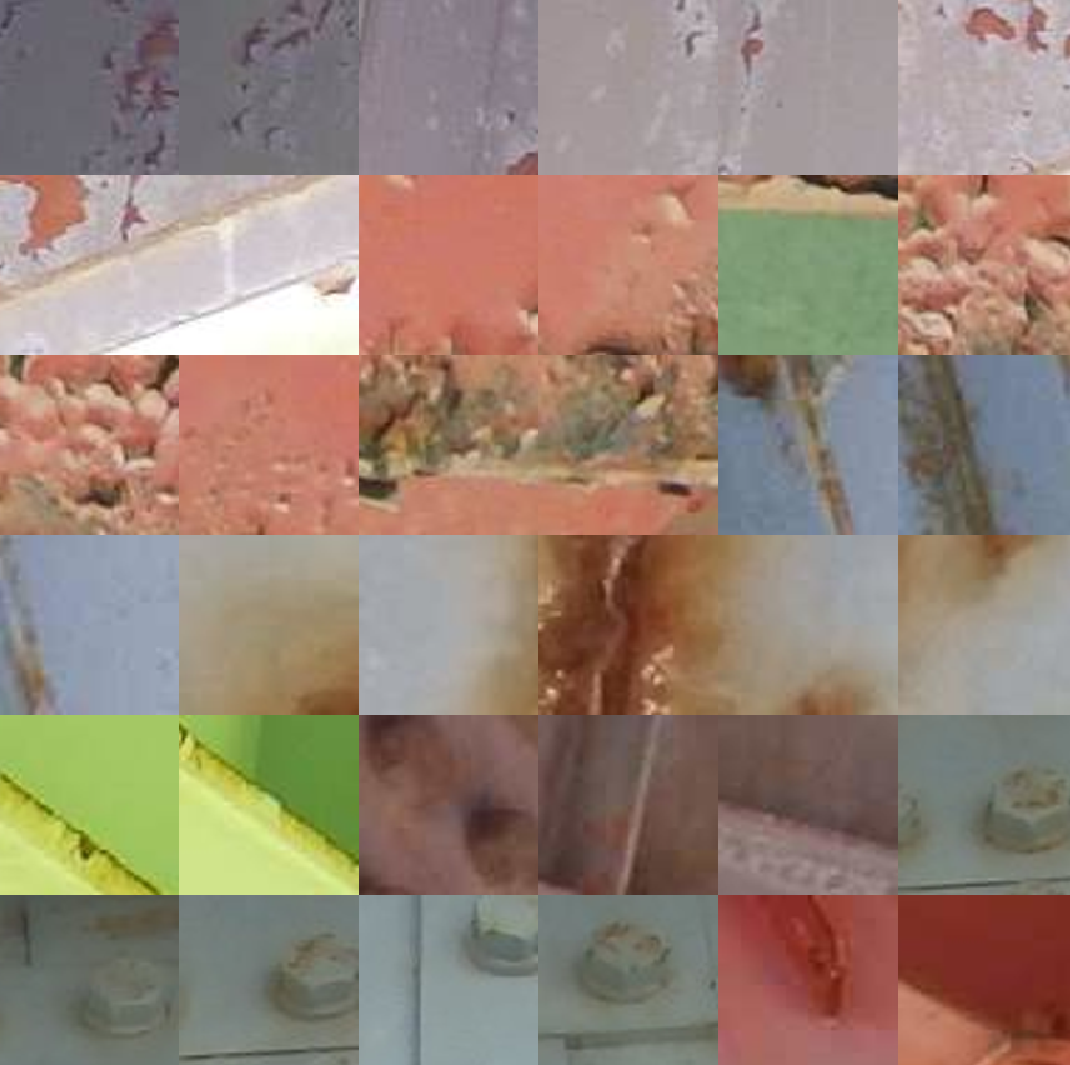}
~\includegraphics[width=0.28\textwidth]{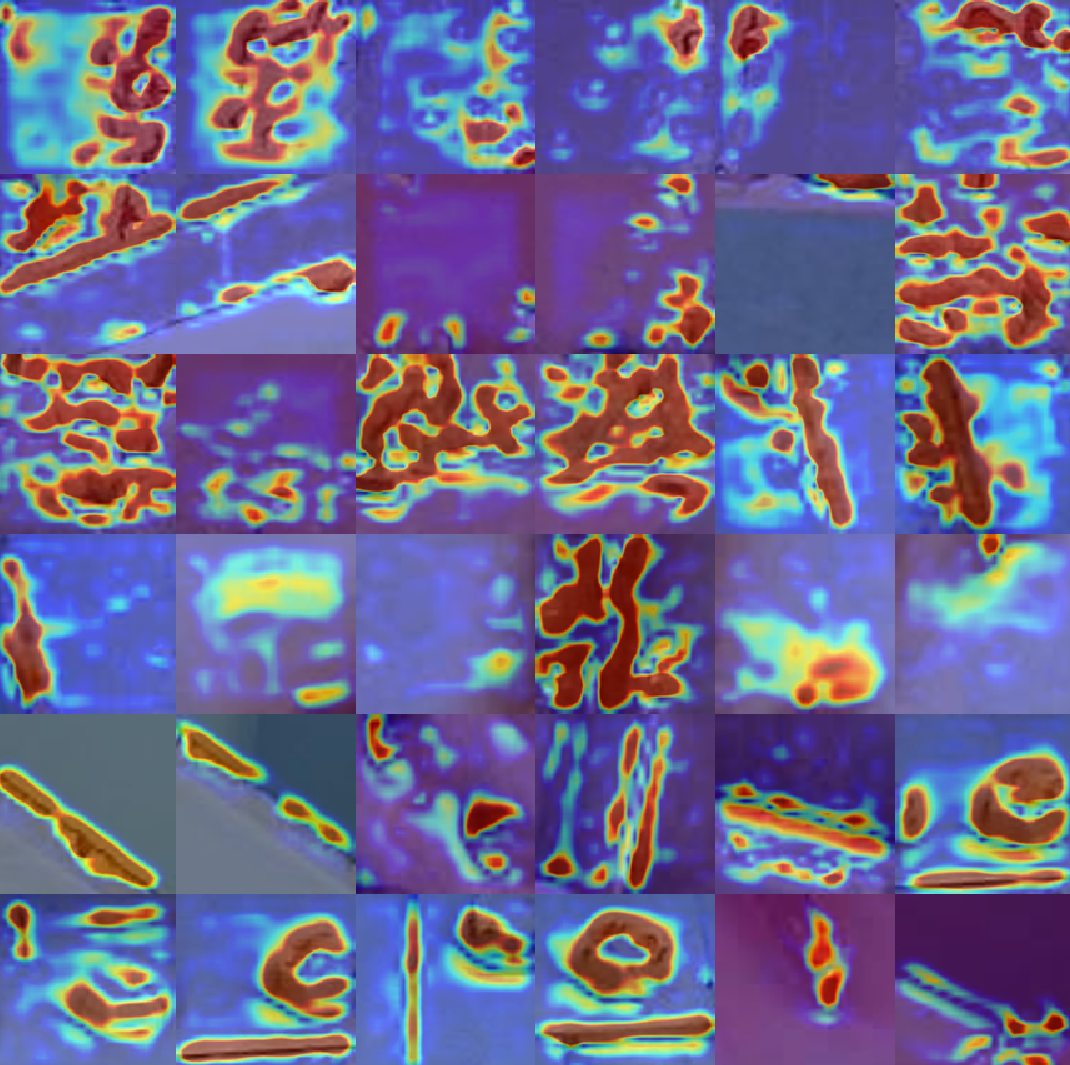}
~\includegraphics[width=0.35\textwidth]{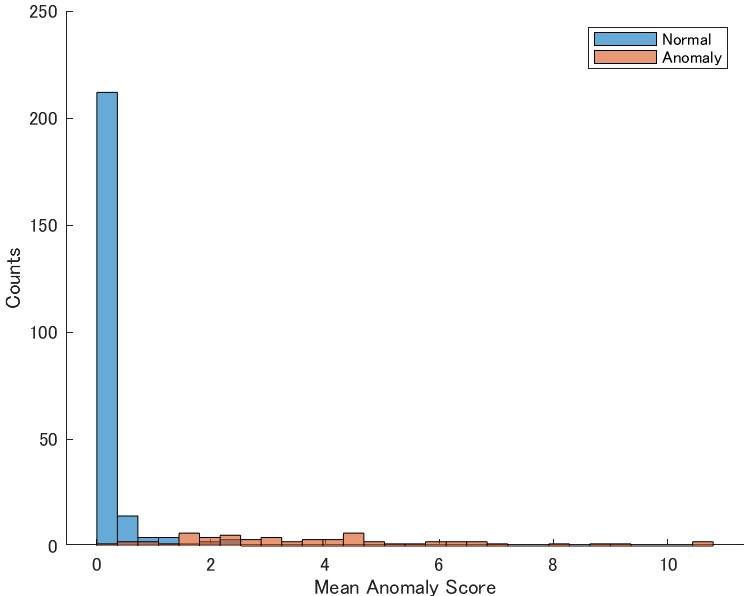}
\caption{\label{fig:rawheatST}Input images (left) of bridge steel corrosion and paint peeling, results for damage mark heatmaps (middle), and a histogram (right) corresponding to the baseline FCDD based on CNN27.}
\end{figure*}
\begin{figure*}[h]
\centering
\includegraphics[width=0.28\textwidth]{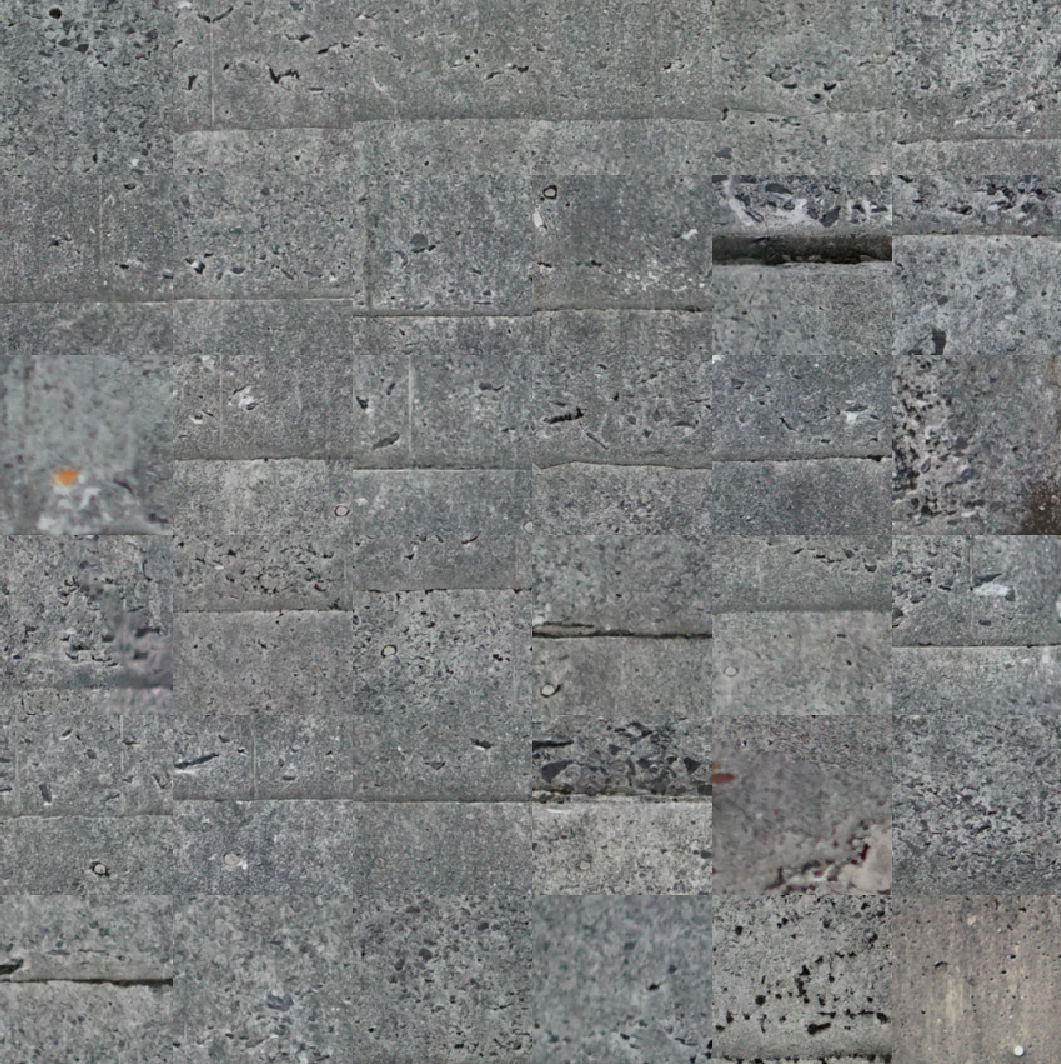}
~\includegraphics[width=0.28\textwidth]{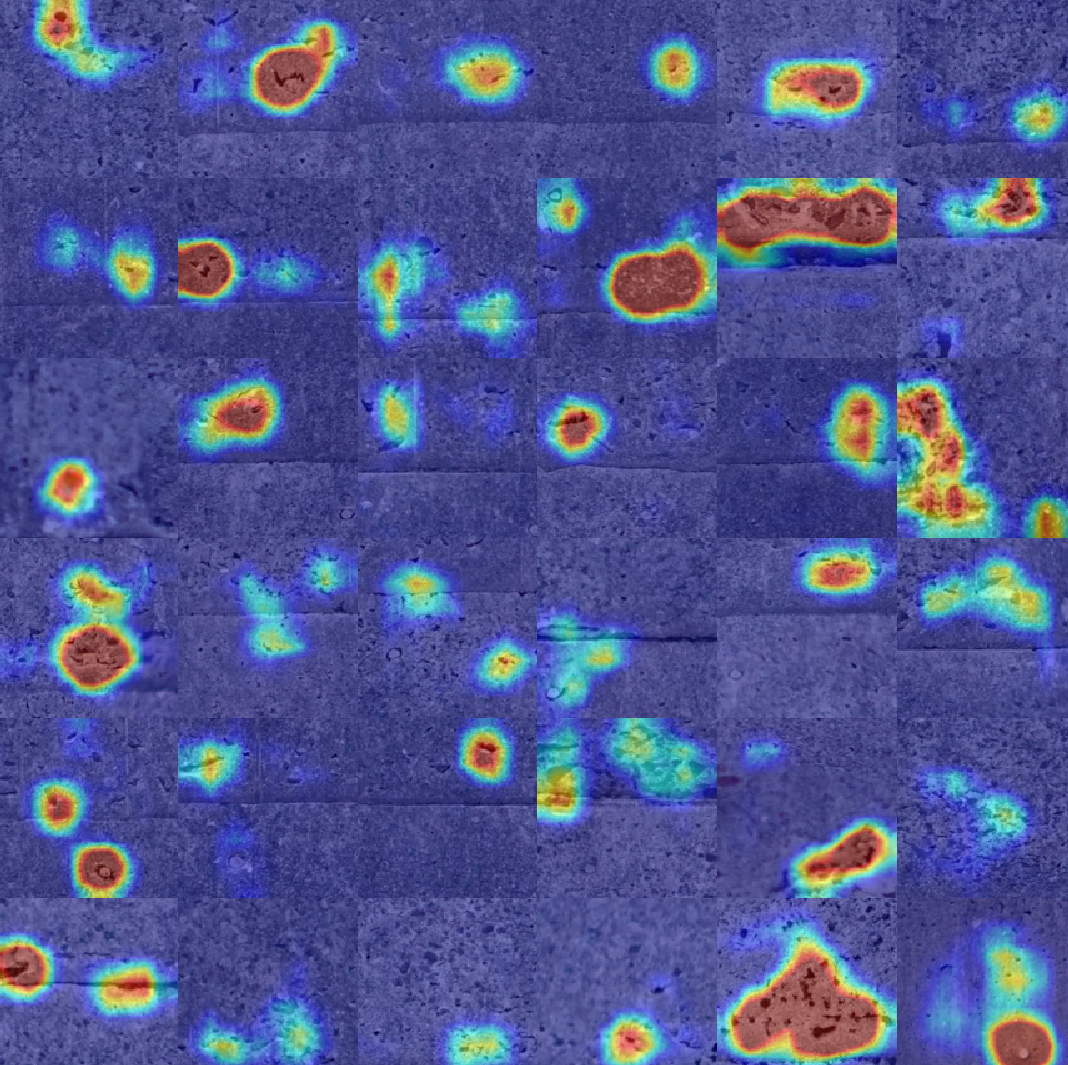}
~\includegraphics[width=0.35\textwidth]{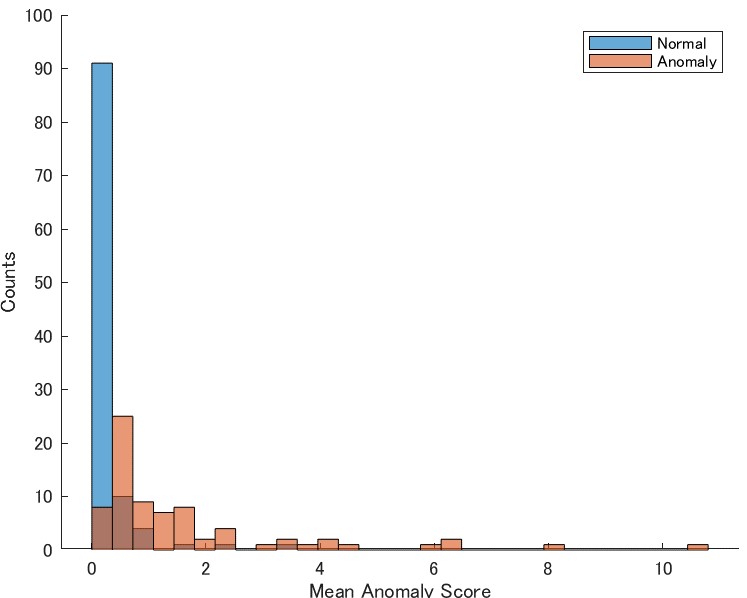}
\caption{\label{fig:rawheatDam}Input images (left) of dam surface janka, results for damage mark heatmaps (middle), and a histogram (right) corresponding to the baseline FCDD based on CNN27.}
\end{figure*}

\subsection{Training the Damage Detector and Accuracy}
The input size was set to $224^2$ while training the damage detector. We set the mini-batch size to 30 and number of epochs to 50. We used the Adam optimizer with a learning rate of 0.0001, set the gradient decay factor to 0.9, and set the squared gradient decay factor to 0.99. The training images were partitioned to set a ratio of 7:1:2 for the numbers of training, calibration, and testing images, respectively. 

\subsection{Civil Damage Mark Heatmaps}
We visualized damage features using the Gaussian upsampling of the receptive field of our CNN27 network. We also generated a histogram of the anomaly scores of test images for four civil engineering datasets. 
First, the middle of Figure \ref{fig:rawheatPave} shows how each heatmap facilitates the visualization of the crack regions of interest to achieve damage mark explanation. Figure \ref{fig:rawheatPave} reveals that three overlapping bins of horizontal anomaly scores exist as a result of shadows in images. 
Second, the middle of Figure \ref{fig:rawheatRebar} shows how each heatmap facilitates the visualization of rebar exposure with either large or small regions to achieve damage mark explanation. The right side of Figure \ref{fig:rawheatRebar} reveals that few overlapping bins exist in the horizontal anomaly scores. Therefore, the score range is well separated for rebar exposure detection.
Third, the middle of Figure \ref{fig:rawheatST} shows how each heatmap facilitates the visualization of paint peeling and volt nut corrosion to achieve damage mark explanation. The right side of Figure \ref{fig:rawheatST} reveals that few overlapping bins exist in the horizontal anomaly scores. Therefore, the score range is well separated for steel paint peeling and corrosion detection.
Finally, the middle of Figure \ref{fig:rawheatDam} shows how each heatmap facilitates the visualization of janka on the surface of the dam embankment to achieve damage mark explanation. The right side of Figure \ref{fig:rawheatDam} reveals that three overlapping bins exist in the horizontal anomaly scores because separating janka features from healthy concrete is visually difficult.

\begin{table*}[h]
\caption{\label{tab:datadisaster}Damage datasets for hurricanes, typhoons, earthquakes, and four-event disasters.}
\centering
\begin{tabular}{c|c|r|r} 
Dataset & Patch size & Normal & Anomalous \\\hline
Hurricane satellite imagery, flooding & $128^2$ & 5,000 & 5,000 \\
Typhoon aerial photography, fallen trees & $486\times442$ & 602 & 698 \\
Earthquake panoramic, building collapse & $224\times256$ & 400 & 400 \\
Disaster drone, four events & $720\times1280,~360\times399$ & 4,390 & 485 \\
\end{tabular}
\end{table*}
\begin{table*}[h]
\caption{\label{tab:accDisaster}Backbone ablation studies on disaster detection using our proposed deeper FCDDs for hurricanes, typhoons, and earthquakes.}
\centering
\begin{tabular}{c|c|c|c|c|c|c}
Dataset & Model & Backbone & AUC & $F_1$ & Precision & Recall \\\hline
                             & baseline FCDD & CNN27 & 0.9892 & 0.9518 & 0.9556 & 0.9480 \\
Hurricane (satellite), & deeper FCDD   & VGG16 & 0.9954 & 0.9781 & 0.9851 & 0.9713 \\
flooding                  & \textbf{deeper FCDD}   &\textbf{ResNet101} &\textbf{0.9982} & \textbf{0.9856} & 0.9879 & \textbf{0.9833} \\
                            & deeper FCDD   &Inceptionv3 &0.9965 & 0.9812 & \textbf{0.9885} & 0.9740 \\ \hline
                           & baseline FCDD & CNN27 & 0.9051 & 0.8104 & 0.8384 & 0.7841 \\
Typhoon (aerial),    & deeper FCDD & VGG16 & 0.9733 & 0.8793 & \textbf{0.9471} & 0.8206 \\
fallen trees           & deeper FCDD & ResNet101 &\textbf{0.9771} & 0.8315 & 0.9420 & 0.7442 \\
                          & \textbf{deeper FCDD} & \textbf{Inceptionv3} &0.9672 & \textbf{0.9047} & 0.9421 & \textbf{0.8702} \\ \hline
                          & baseline FCDD & CNN27 &0.9987 & 0.9816 & 0.9638 & 1.000 \\
Earthquake (panoramic),& deeper FCDD & VGG16 &0.9962 & 0.9916 & 0.9916 & 0.9916 \\
building collapse   & deeper FCDD & ResNet101 &0.9987 & 0.9958 & 1.000 & 0.9916 \\
                         & \textbf{deeper FCDD} & \textbf{Inceptionv3} &\textbf{1.000} & \textbf{0.9958} & \textbf{1.000} & \textbf{0.9916} \\ \hline
                         & baseline FCDD & CNN27 &0.9433 & 0.7896 & 0.7523 & 0.8307 \\
Disaster (drone),  & \textbf{deeper FCDD} & \textbf{VGG16} &\textbf{0.9969} & \textbf{0.9622} & \textbf{0.9589} & 0.9655 \\
four events         & deeper FCDD & ResNet101 &0.9916 & 0.9323 & 0.8985 & \textbf{0.9687} \\
                         & deeper FCDD & Inceptionv3 &0.9925 & 0.9319 & 0.9189 & 0.9453 \\ 
\end{tabular}
\end{table*}

\section{Ablation Studies using Deeper FCDDs}

\subsection{Damage Datasets from Natural Disasters}
To develop a robust application, we evaluated our method on datasets containing images of natural disaster damage caused by hurricanes \cite{texas2018}, typhoons \cite{asahi2020}, earthquakes \cite{sakurada2015}, and combinations of multiple disasters, including collapsed buildings, traffic incidents, fires, and floods \cite{AIDER2019}.
These disaster images were collected using various modes, including satellite imagery, aerial photography, drone-based systems, and panoramic 360 cameras.  
As shown in Table~\ref{tab:datadisaster}, we also evaluated our method via ablation studies using deeper backbones, namely VGG16, ResNet101, and Inceptionv3.

Regarding the dataset used for ablation studies, the hurricane dataset \cite{texas2018} consisted of satellite images from Texas following Hurricane Harvey, which were divided into two groups: damage and no damage. This hurricane caused landfall in Texas and Louisiana in August of 2017, causing devastating flooding and multiple deaths.
The typhoon dataset \cite{asahi2020} was an aerial photography dataset containing images with dimensions of 14000 × 15000 pixels recorded in the South Chiba region 18 days after the typhoon disaster that occurred on September 27 and 28 of 2019. This dataset was provided by Aero Asahi Co. Ltd. The real land dimension per pixel was 19.6 cm; therefore, each unit grid square covered an area of 44 × 48 m2.
The earthquake dataset \cite{sakurada2015} was created as a panoramic change-detection dataset for experiments. This dataset contained 100 panoramic image pairs of scenes from tsunami-damaged areas in Japan from March of 2011. The size of the panoramic images was 224$\times$1024 pixels.
The multiple-disaster dataset \cite{AIDER2019} was an aerial image dataset developed for emergency response applications. The construction of the dataset involved manually collecting images for four types of disaster events, namely fire/smoke, flood, collapsed building/rubble, and traffic accidents, as well as an additional class for the normal state.

\subsection{Training the Disaster Detector with Deeper Backbones}
Initially, we trained a baseline FCDD with the aforementioned backbone CNN27, which had neither a skip layer nor residual layer. 
Additionally, we constructed FNCs with deeper backbones based on VGG16, ResNet101, and Inceptionv3, which contained either skip, residual, or mixed layers of various scales. 
Table~\ref{tab:accDisaster} presents the accuracy values for one-class damage detection when applying the models to the natural disaster datasets representing hurricanes, typhoons, and earthquakes. The AUC and recall values are high on the natural disaster dataset. 
This suggests that the FCDD could be applied to complex and noisy damage images for disaster detection.
From the perspective of accuracy, in the case of hurricane satellite imagery and typhoon aerial photography, the FCDD model with the ResNet101 backbone outperformed the other models with different backbones. In the case of earthquake drone images, the FCDD model with the VGG16 backbone outperformed the other models. In contrast, in the case of earthquake panoramic camera images, the FCDD model with the Inceptionv3 backbone outperformed the other models. 

\begin{figure*}[h]
\centering
\includegraphics[width=0.28\textwidth]{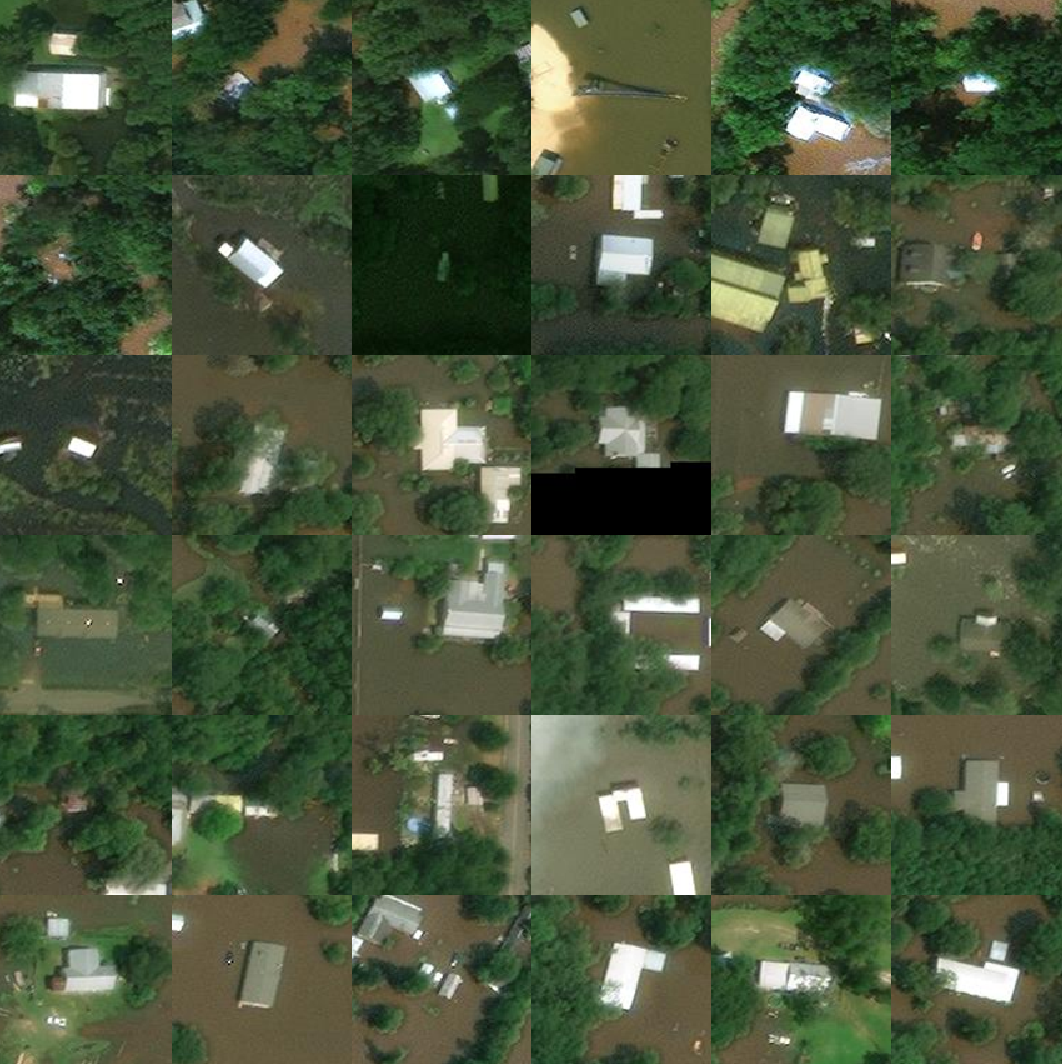}
~\includegraphics[width=0.28\textwidth]{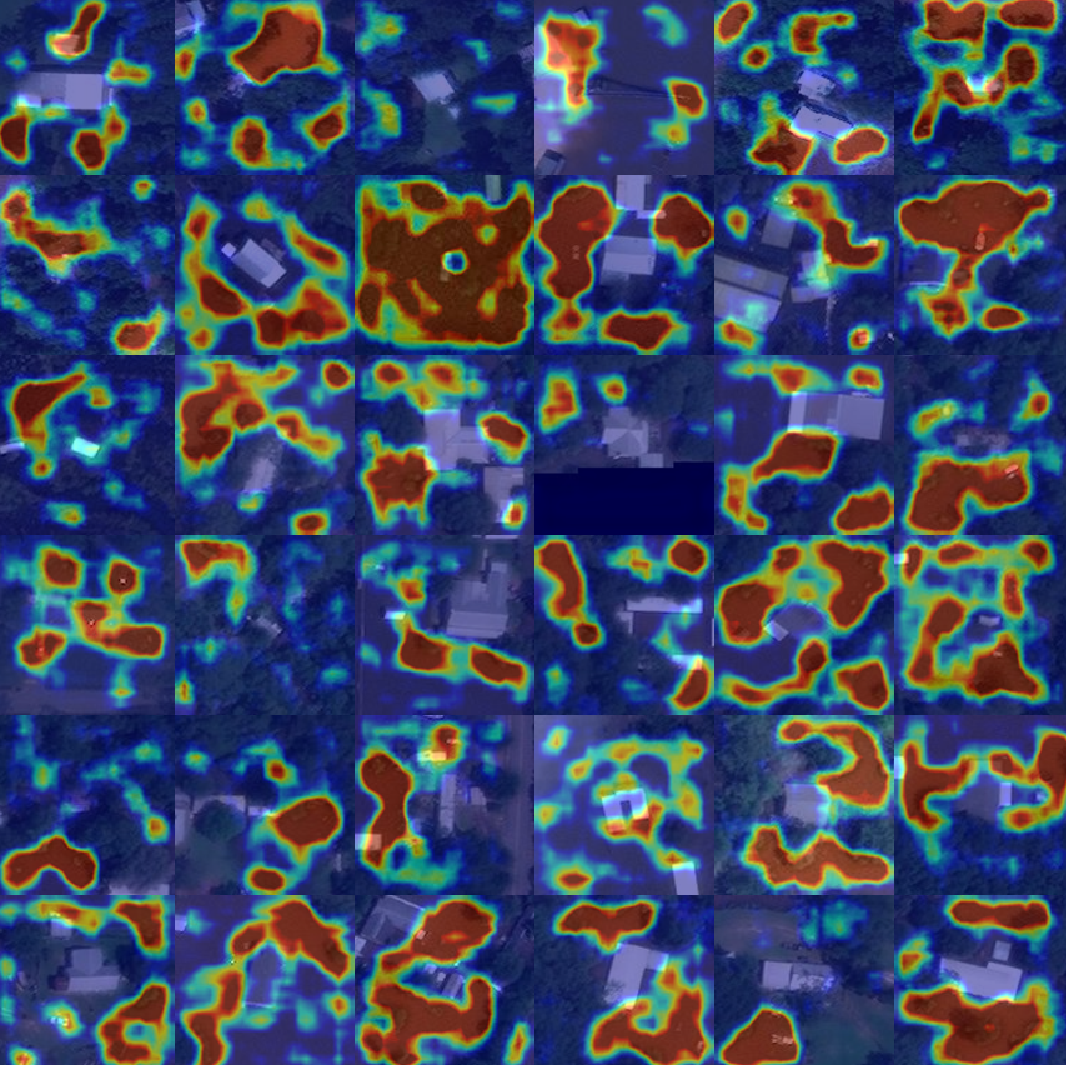}
~\includegraphics[width=0.35\textwidth]{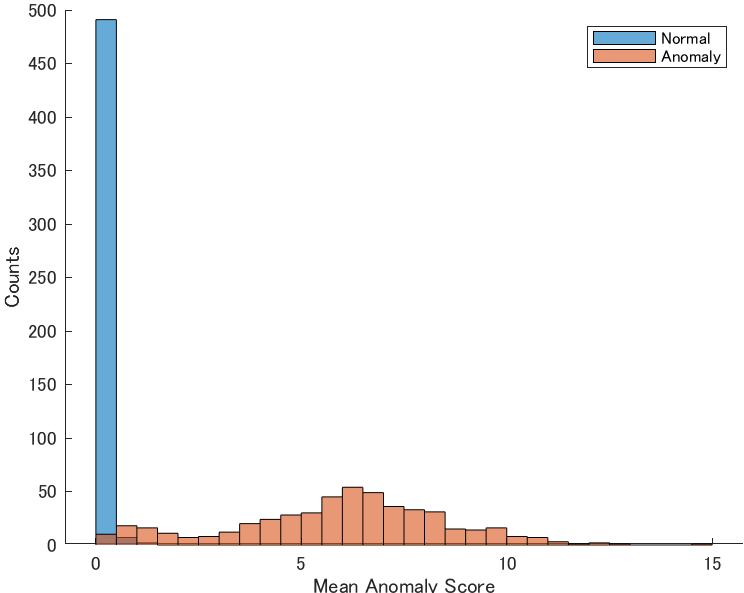}
\caption{\label{fig:rawheatHurri}Input images (left) of hurricane (satellite imagery) and flood damage, results for damage mark heatmaps (middle), and a histogram (right) corresponding to our deeper FCDD based on the ResNet101 backbone.}
\end{figure*}
\begin{figure*}[h]
\centering
\includegraphics[width=0.28\textwidth]{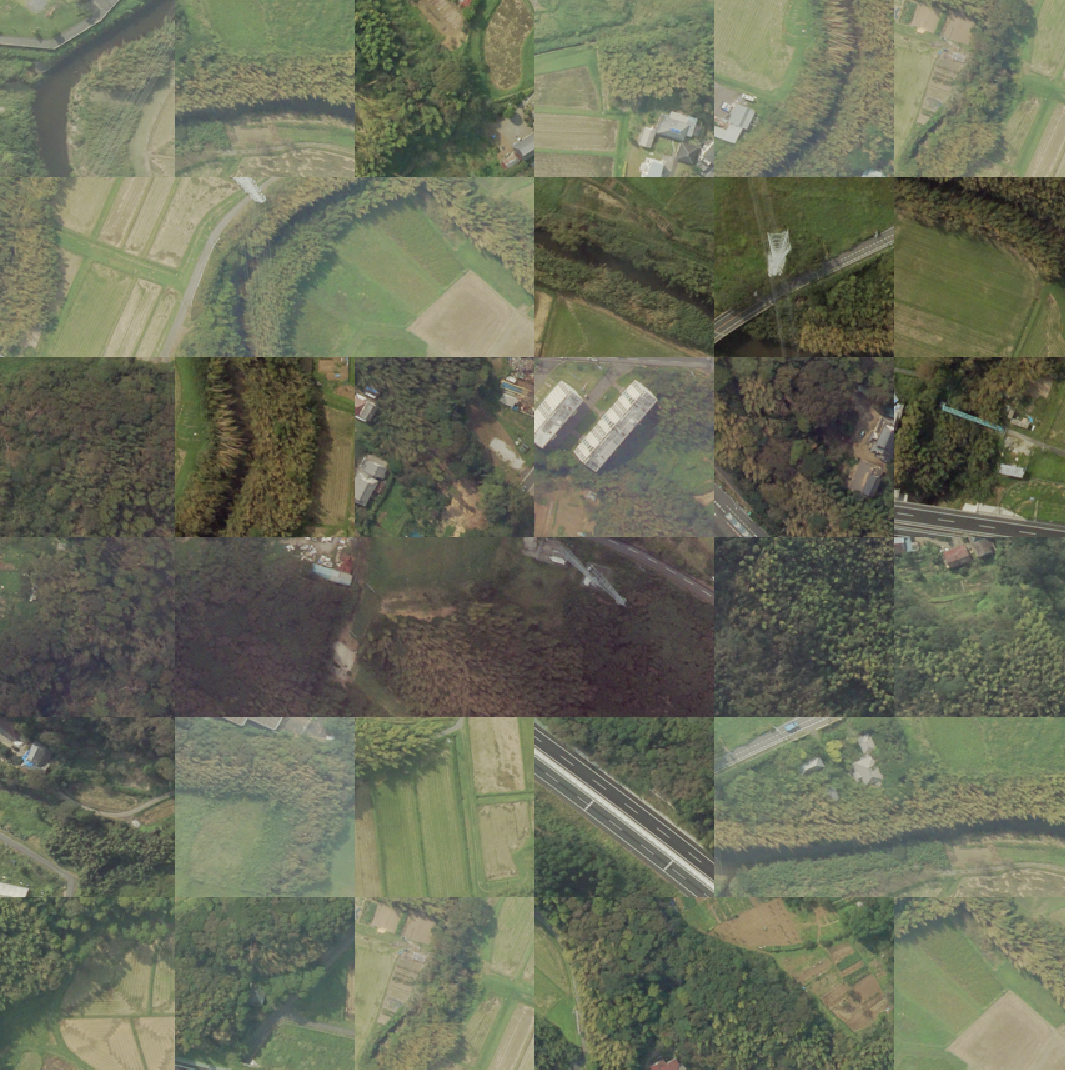}
~\includegraphics[width=0.28\textwidth]{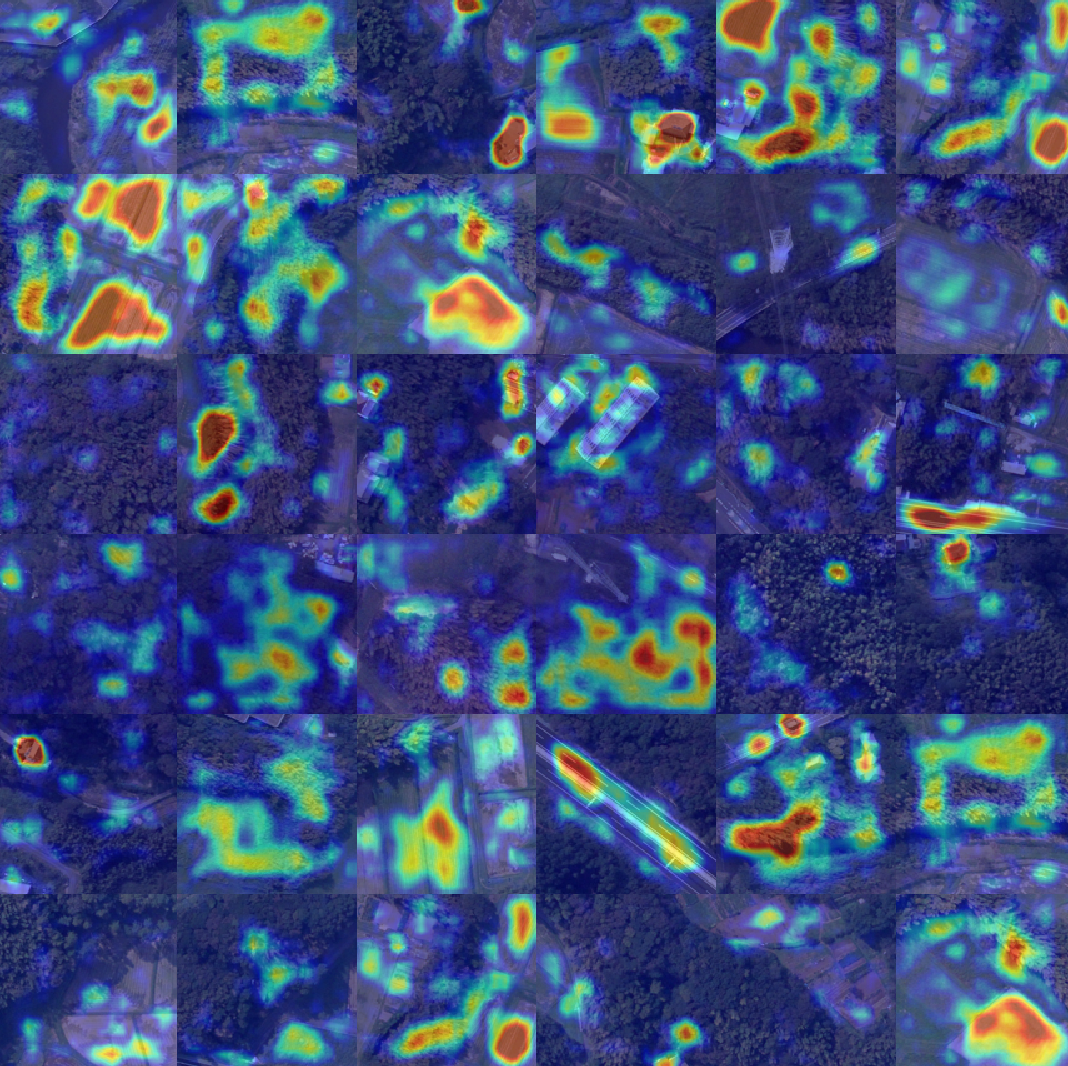}
~\includegraphics[width=0.35\textwidth]{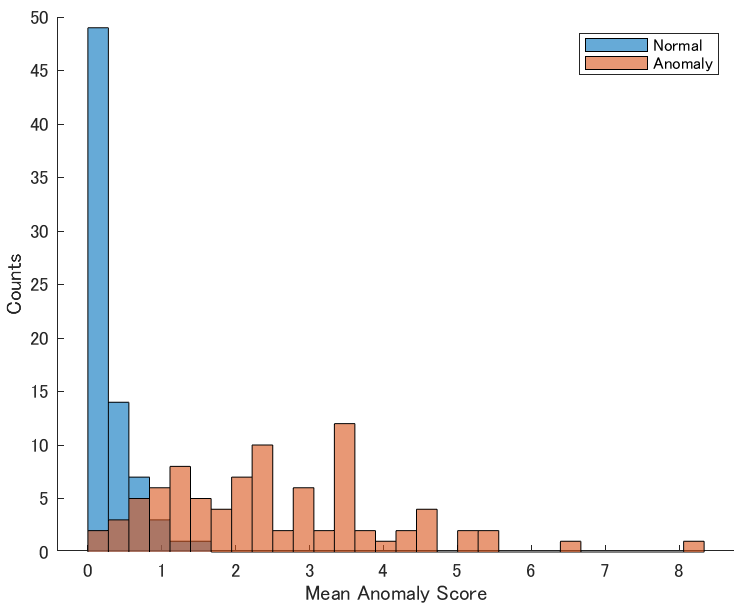}
\caption{\label{fig:rawheatTyphoon}Input images (left) of typhoon (aerial photography) and fallen trees, results for damage mark heatmaps (middle), and a histogram (right) corresponding to our deeper FCDD based on the Inceptionv3 backbone.}
\end{figure*}
\begin{figure*}[h]
\centering
\includegraphics[width=0.28\textwidth]{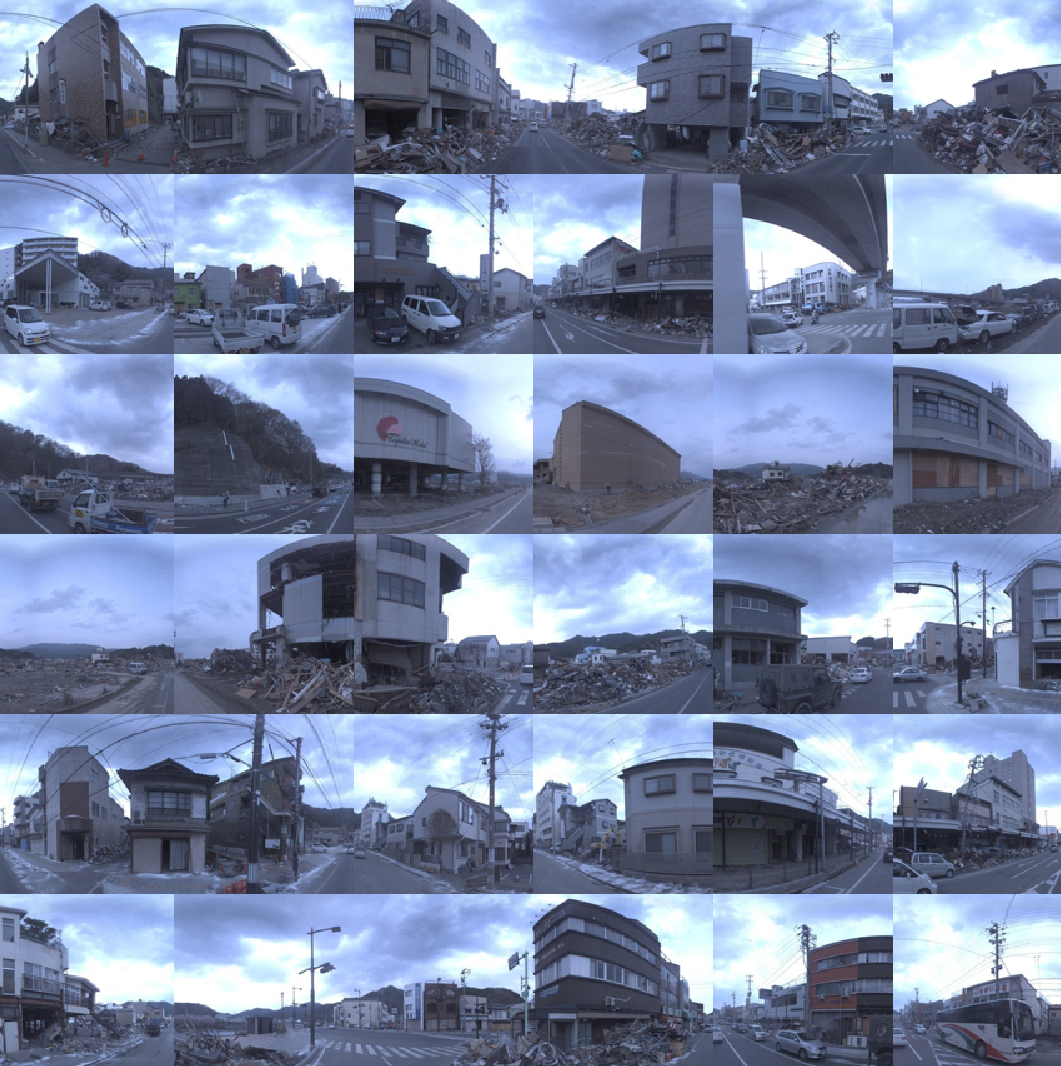}
~\includegraphics[width=0.28\textwidth]{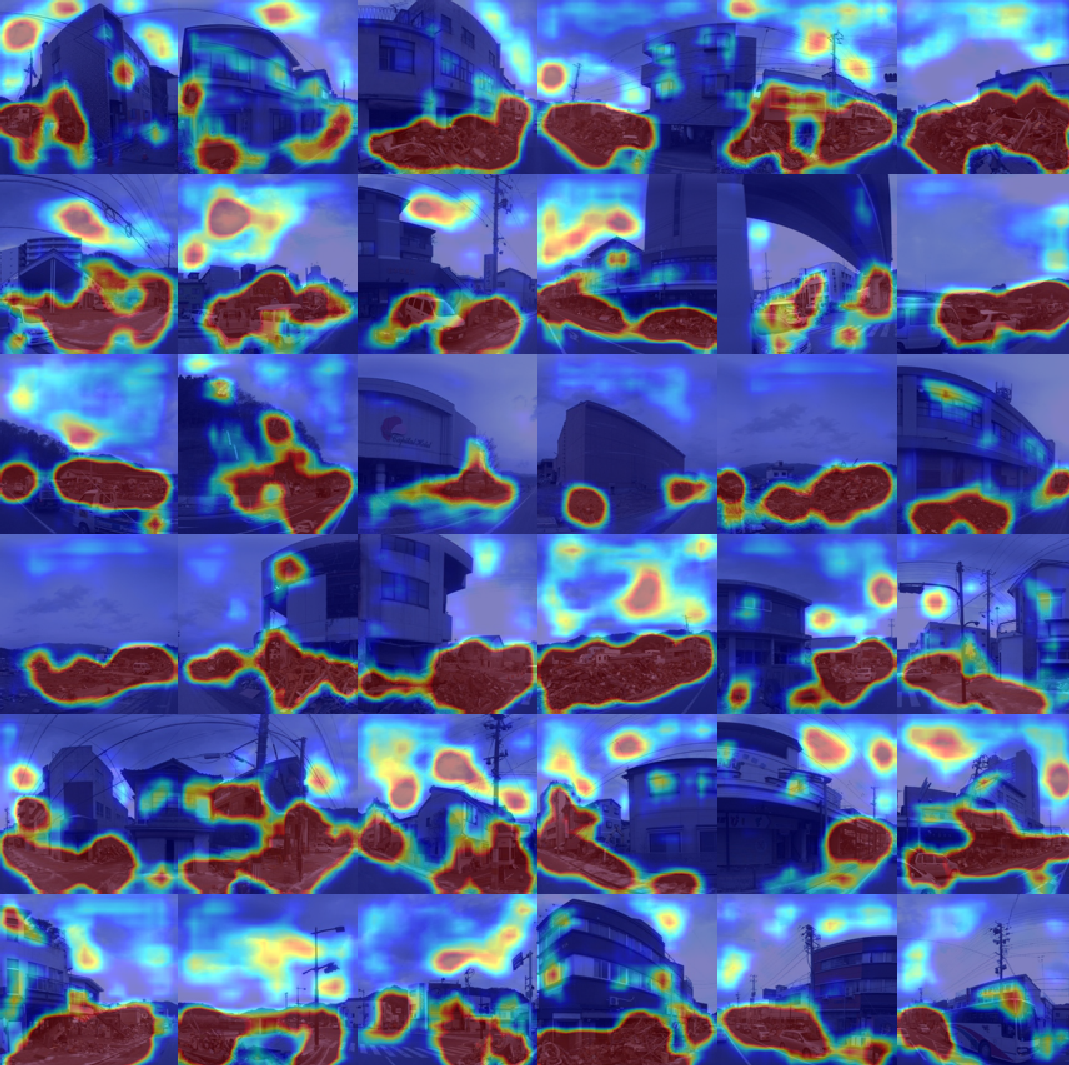}
~\includegraphics[width=0.35\textwidth]{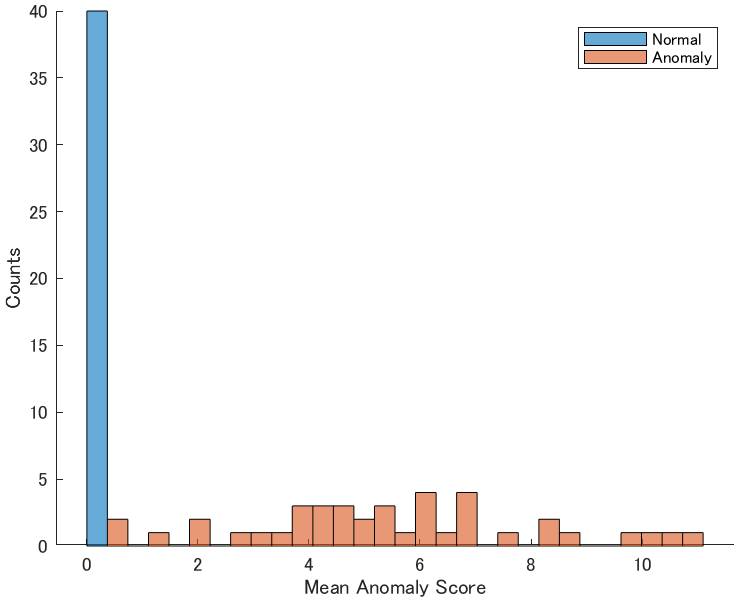}
\caption{\label{fig:rawheatQuake360}Input images (left) of earthquake (panoramic camera) and building collapse, results for damage mark heatmaps (middle), and a histogram (right) corresponding to our deeper FCDD based on the Inceptionv3 backbone.}
\end{figure*}
\begin{figure*}[h]
\centering
\includegraphics[width=0.28\textwidth]{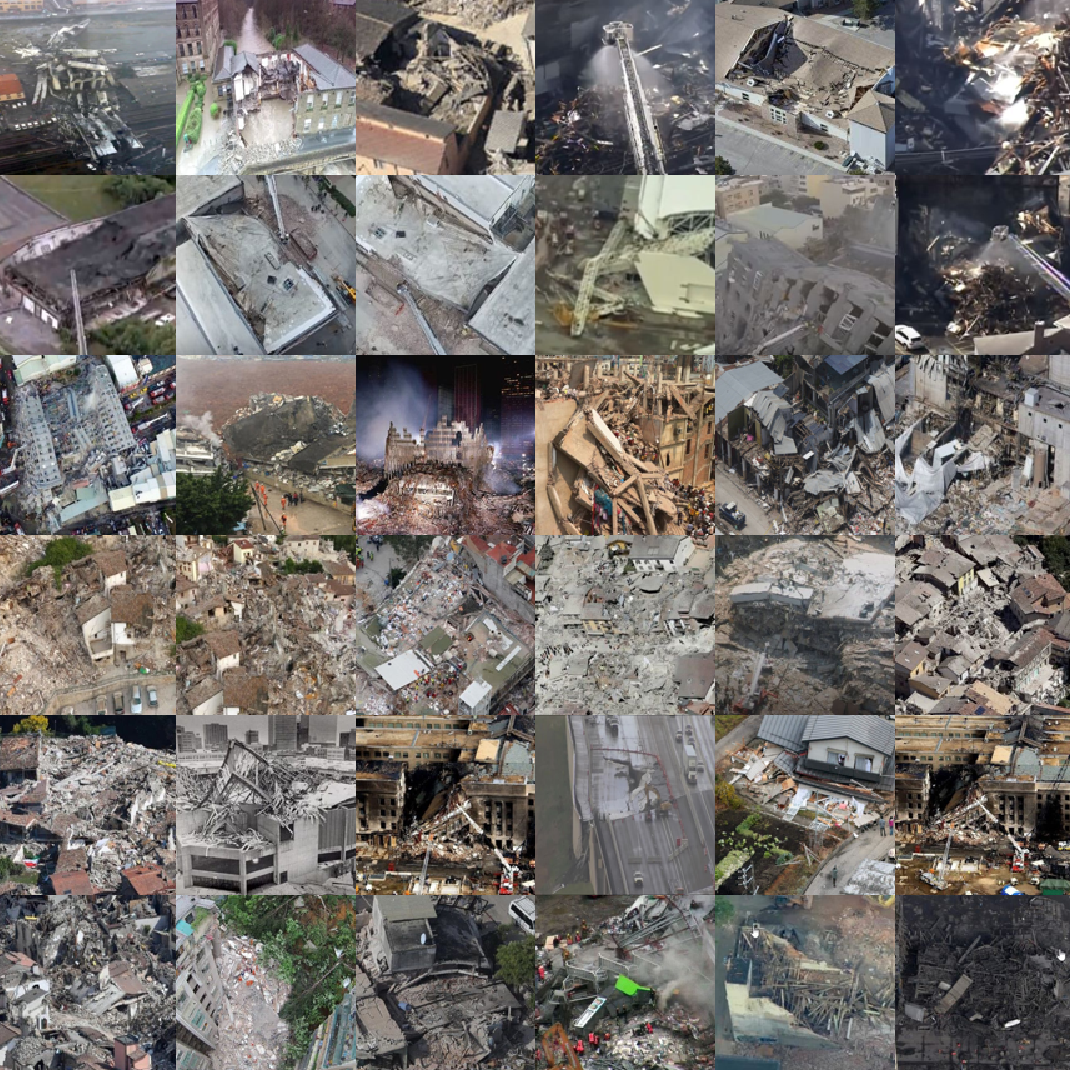}
~\includegraphics[width=0.28\textwidth]{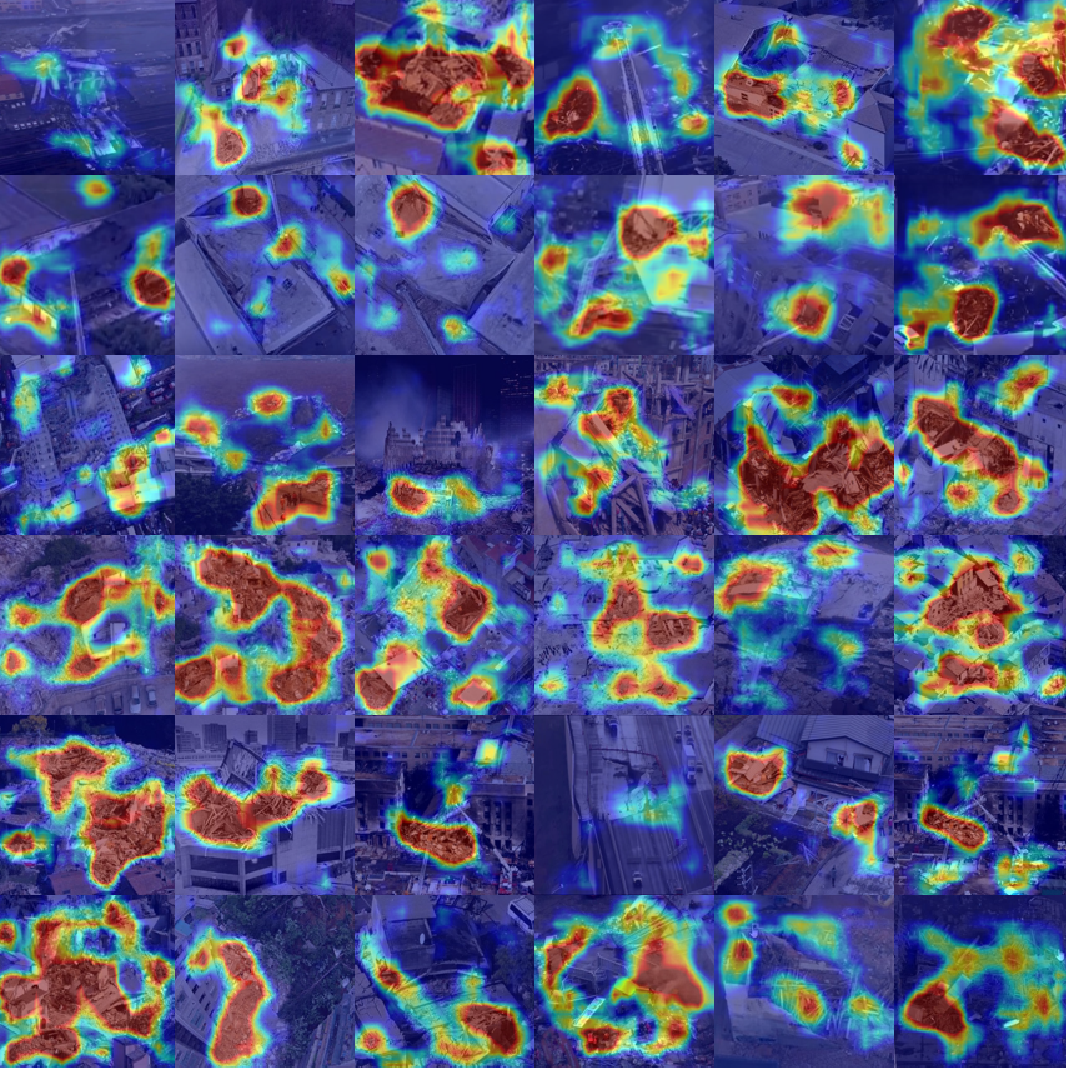}
~\includegraphics[width=0.35\textwidth]{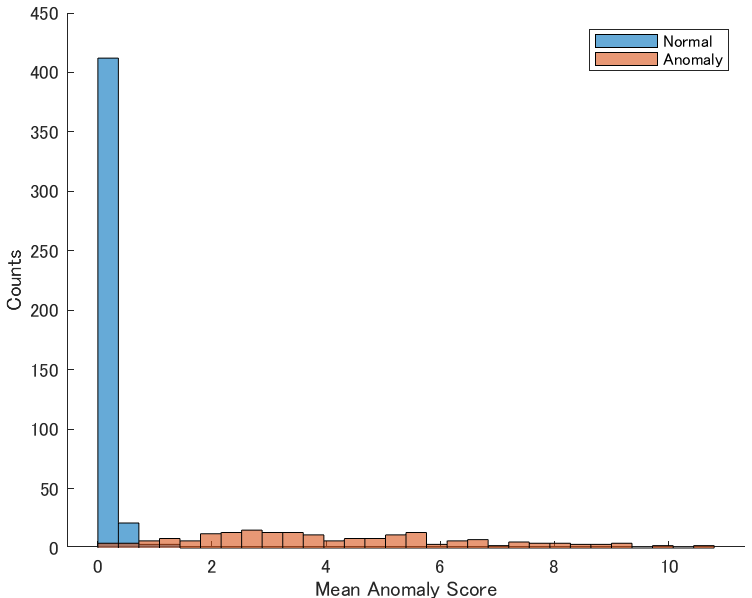}
\caption{\label{fig:rawheatDisasterDrone}Input images (left) of collapsed buildings among four-event disasters (drone), results for damage mark heatmaps (middle), and a histogram (right) corresponding to our deeper FCDD based on the VGG16 backbone.}
\end{figure*}

\subsection{Disaster Damage Mark Heatmaps}
First, the middle of Figure \ref{fig:rawheatHurri} shows how each heatmap facilitates visualization of flooding areas of interest to achieve damage mark explanation. Figure \ref{fig:rawheatHurri} reveals that few overlapping bins exist in the horizontal anomaly scores. Therefore, the score range is well separated for flood damage detection. 
Second, the middle of Figure \ref{fig:rawheatTyphoon} shows how each heatmap facilitates the visualization of the fallen tree regions to achieve damage mark explanation. The right side of Figure \ref{fig:rawheatTyphoon} reveals that three overlapping bins exist in the horizontal anomaly scores because separating fallen tree features is difficult.
Third, the middle of Figure \ref{fig:rawheatQuake360} shows how each heatmap facilitates the visualization of construction waste to achieve damage mark explanation. The right side of Figure \ref{fig:rawheatQuake360} reveals that no overlapping bins exist in the horizontal anomaly scores. Therefore, the score range is well separated for construction waste detection following tsunami damage.
Finally, the middle of Figure \ref{fig:rawheatDisasterDrone} shows how each heatmap facilitates the visualization of collapsed buildings to achieve damage mark explanation. The right side of Figure \ref{fig:rawheatDisasterDrone} reveals that few overlapping bins exist in the horizontal anomaly scores. Therefore, the score range is well separated for disaster detection.

\section{Concluding Remarks}

\subsection{Robust Damage Detection for Civil and Disaster Applications}
We constructed a civil-purpose application to automate one-class damage detection reproducing a baseline FCDD with a light backbone CNN network containing 27 layers with either Conv-BN-ReLU or Maxpooling activation. 
We also visualized damage mark heatmaps using direct Gaussian upsampling of the receptive field of the FCN. We evaluated the baseline FCDD model on four experimental targets, namely concrete pavement cracks, rebar exposure on bridge components, steel corrosion, and dam embankment janka. Our experiments yielded high accuracy of AUC and recall. Therefore, the lightweight FCDD may be applicable for infrastructure damage inspection. Without annotating damage regions, the FCDD enhanced damage marks for visual explanation. 
To develop a more robust application, we evaluated a novel solution of deeper FCDDs with pre-trained backbones of VGG16, ResNet101, and Inceptionv3, and performed ablation studies via comparisons with a baseline FCDD. We applied our model to datasets representing natural disaster damage caused by including hurricanes, typhoons, earthquakes, and four-event disasters.
We have found that a robust solution of deeper FCDDs outperformed the baseline FCDD on these complex datasets.
A novel solution of deeper FCDDs provides a powerful tool for damage vision applications utilized in the high accuracy, explainability, and robustness.

\subsection{Future Works}
Several promising directions exist for future works to develop more accurate and robust applications. 
For more robust training in the presence of background noise, an augmentation preprocessing operation could be effective for one-class classification models. Such operations include mixup, RICAP, cutout, and random erasing. To achieve unified applicability, a unified framework could be constructed, wherein the data domain of each dataset is pre-classified to guide data classification.
Following data domain classification, damage features could be detected using deeper FCDDs. 
For efficient data mining, a damage detector based on FCDDs could be used at edge devices such as IP cameras, drones, aerial photography platforms, and satellites. Instead of collecting all image files, only damage-marked images that have a significantly higher score than a predefined threshold could be efficiently collected. FCDDs require less memory for training a damage detector and computing an upsampling heatmap.

\subsection*{Acknowledgments} We gratefully acknowledge the conductive comments of the anonymous referees. The authors wish to thank MathWorks and Takuji Fukumoto, who provided helpful MATLAB resources for automated visual inspection using anomaly detection.


\bibliography{ISARC}

\begin{thebibliography}{42}
\providecommand{\natexlab}[1]{#1}
\providecommand{\url}[1]{\texttt{#1}}
\expandafter\ifx\csname urlstyle\endcsname\relax
  \providecommand{\doi}[1]{doi: #1}\else
  \providecommand{\doi}{doi: \begingroup \urlstyle{rm}\Url}\fi

\bibitem[V.~Chandola(2009)]{Chandola2009}
V.~Kumar V.~Chandola, A.~Banerjee.
\newblock Anomaly detection : A survey.
\newblock \emph{ACM Computing Surveys}, 09:\penalty0 1--72, 2009.

\bibitem[R.~Chalapathy(2019)]{Chalapathy2019}
S.~Chawla R.~Chalapathy.
\newblock Deep learning for anomaly detection : A survey, 2019.

\bibitem[L.Ruff(2020)]{Ruff2020}
R.A.~Vandermeulen L.Ruff, J.R.~Kauffmann.
\newblock A unifying review of deep and shallow anomaly detection, 2020.

\bibitem[S.~Yuan(2022)]{Yuan2022}
X.~Wu S.~Yuan.
\newblock Trustworthy anomaly detection : A survey, 2022.

\bibitem[A.~Saberironaghi and El-Gindy(2023)]{Saberironaghi2023}
J.~Ren A.~Saberironaghi and M.~El-Gindy.
\newblock Defect detection methods for industrial products using deep learning
  techniques: A review.
\newblock \emph{Algorithms}, 16-95, 2023.

\bibitem[Payawal and Kim(2023)]{Payawal2023}
J.~M.~G. Payawal and D.~K. Kim.
\newblock Image-based structural health monitoring: A systematic review.
\newblock \emph{Applied Science}, 13-968, 2023.

\bibitem[Xu~S.(2019)]{Wang2019}
Shou~W. Xu~S., Wang~J.
\newblock Computer vision technique in construction, operation and maintenance
  phases of civil assets: A critical review.
\newblock In \emph{36th International Symposium on Automation and Robotics in
  Construction (ISARC)}, 2019.

\bibitem[E.~Bianchi(2022)]{Bianchi2022}
M.~Hebdon E.~Bianchi.
\newblock Visual structural inspection datasets.
\newblock \emph{Automation in Construction}, 139(2):104299, 2022.

\bibitem[R.~Chalapathy and Chawla(2018)]{Chalapathy2018}
A.~K.~Menon R.~Chalapathy and S.~Chawla.
\newblock Anomaly detection using one-class neural networks, 2018.

\bibitem[Tax and Duin(2004)]{Tax2014}
D.~M.~J. Tax and R.~P. Duin.
\newblock Support vector data description.
\newblock \emph{Machine Learning}, 54(1):\penalty0 45--66, 2004.

\bibitem[Hawkins(1974)]{Hawkins1974}
D.~M. Hawkins.
\newblock Detection of errors in multivariate data using principal components.
\newblock \emph{J. of American Statistical Association}, 69-346:\penalty0
  340--344, 1974.

\bibitem[Hoffmann(2007)]{Hoffmann2007}
H.~Hoffmann.
\newblock Kernel pca for novelty detection.
\newblock \emph{Pattern Recognition}, 40:\penalty0 863--874, 2007.

\bibitem[L.~Ruff and Kloft(2018)]{Ruff2018}
N.~Gornitz L. Deecke S. A. Siddiqui A. Binder E.~Muller L.~Ruff, R.
  A.~Vandermeulen and M.~Kloft.
\newblock Deep one-class classification.
\newblock \emph{International Conference on Machine Learning}, 80:\penalty0
  4390--4399, 2018.

\bibitem[P.~Liznerski(2021)]{Liznerski2021}
et~al. P.~Liznerski, L.~Ruff.
\newblock Explainable deep one-class classification.
\newblock \emph{The International Conference on Learning
  Representations(ICLR)}, 2021.

\bibitem[Kingma and Welling(2019)]{Kingma2019}
D.~P. Kingma and M.~Welling.
\newblock An introduction to variational autoencoders.
\newblock \emph{Foundations and Trends in Machine Learning}, 12-4:\penalty0
  307--392, 2019.

\bibitem[An and Cho(2015)]{An2015}
J.~An and S.~Cho.
\newblock Variational autoencoder based anomaly detection using reconstruction
  probability.
\newblock \emph{Special Lecture on IE}, 2:\penalty0 1--18, 2015.

\bibitem[Zhou and Paffenroth(2017)]{Zhou2017}
C.~Zhou and R.~C. Paffenroth.
\newblock Anomaly detection with robust deep autoencoders.
\newblock \emph{International Conference on Knowledge Discovery and Data
  Mining}, pages 665--674, 2017.

\bibitem[R.~Chan and for Automated~Driving(2022)]{Chan2022}
M.~Rottmann H. Gottschalk In: Fingscheidt T. Gottschalk H. Houben S. (eds) Deep
  Neural~Networks R.~Chan, S.~Uhlemeyer and Data for Automated~Driving.
\newblock \emph{Detecting and Learning the Unknown in Semantic Segmentation}.
\newblock Springer, 2022.

\bibitem[D.~Hendrycks(2017)]{Hendrycks2017}
K.~Gimpel D.~Hendrycks.
\newblock A baseline for detecting misclassified and out-of-distribution
  examples in neural networks.
\newblock In \emph{Proceedings of the International Conference on Learning
  Representations (ICLR)}, pages 1--12, 2017.

\bibitem[S.~Liang(2018)]{Liang2018}
R.~Srikant S.~Liang, Y.~Li.
\newblock Enhancing the reliability of out-of-distribution image detection in
  neural networks.
\newblock In \emph{Proceedings of the International Conference on Learning
  Representations (ICLR)}, pages 1--15, 2018.

\bibitem[G.~Di~Biase(2021)]{Biase2021}
R.~Siegwart C.~Cadena G.~Di~Biase, H.~Blum.
\newblock Pixel-wise anomaly detection in complex driving scenes.
\newblock In \emph{Proceedings of the IEEE/CVF Conference on Computer Vision
  and Pattern Recognition}, pages 16918–--16927, 2021.

\bibitem[R.~Chan(2021)]{Chan2021}
H.~Gottschalk R.~Chan, M.~Rottmann.
\newblock Entropy maximization and meta classification for out-of-distribution
  detection in semantic segmentation.
\newblock In \emph{Proceedings of the IEEE International Conference on Computer
  Vision (ICCV)}, pages 5128--–5137, 2021.

\bibitem[D.~Fontanel(2021)]{Fontanel2021}
M.~Mancini B.~Caputo D.~Fontanel, F.~Cermelli.
\newblock Detecting anomalies in semantic segmentation with prototypes.
\newblock In \emph{CVF Conference on Computer Vision and Pattern Recognition,
  CVPR workshop}, 2021.

\bibitem[Cohen and Hoshen(2020)]{Cohen2020}
N.~Cohen and Y.~Hoshen.
\newblock Sub-image anomaly detection with deep pyramid correspondences, 2020.

\bibitem[D.~Thomas(2020)]{Thomas2020}
L.~Angelique A.~Romaric D.~Thomas, S.~Aleksandr.
\newblock Padim: a patch distribution modeling framework for anomaly detection
  and localization, 2020.

\bibitem[K.~Roth(2021)]{Roth2021}
J.~Zepeda B. Scholkopf T. Brox P.~Gehler K.~Roth, L.~Pemula.
\newblock Towards total recall in industrial anomaly detection.
\newblock In \emph{CVF Conference on Computer Vision and Pattern Recognition},
  2021.

\bibitem[J.~Yu(2021)]{Yu2021}
X.~Wang W. Li Y. Wu R. Zhao L.~Wu J.~Yu, Y.~Zheng.
\newblock Fastflow: Unsupervised anomaly detection and localization via 2d
  normalizing flows, 2021.

\bibitem[T.~Yasuno(2021{\natexlab{a}})]{YasunoIWSHM2021}
M.~Nakajima T.~Yasuno, J.~Fujii.
\newblock Bridge slab anomaly detector using u-net generator with patch
  discriminator for robust prognosis.
\newblock In \emph{Structural Health Monitoring, Proceeding of IWSHM},
  2021{\natexlab{a}}.

\bibitem[T.~Yasuno(2020)]{YasunoISARC2020}
J.~Fujii M. Amakata et~al. T.~Yasuno, A.~Ishii.
\newblock Generative damage learning for concrete aging detection using
  auto-flight images.
\newblock \emph{ISARC}, 2020.

\bibitem[T.~Yasuno(2022)]{YasunoJSAI2022}
R.~Ogata M.~Okano T.~Yasuno, J.~Fujii.
\newblock Vae-iforest: Auto-encoding reconstruction and isolation-based
  anomalies detecting fallen objects on road surface.
\newblock In \emph{JSAI}, 2022.

\bibitem[T(2020)]{asahi2020}
M.~Amakata M.~Okano T, Yasuno.
\newblock Natural disaster classification using aerial photography explainable
  for typhoon damaged feature.
\newblock In \emph{MAES Workshop, ICPR2020}, 2020.

\bibitem[T.~Yasuno(2021{\natexlab{b}})]{YasunoAAI2021}
J.~Fujii T.~Yasuno, H.~Sugawara.
\newblock Road surface translation under snow-covered and semantic segmentation
  for snow hazard index.
\newblock \emph{Advances in Artificial intelligence - Selected Papers from the
  Annual Conference of JSAI2021}, 2021{\natexlab{b}}.

\bibitem[T.~Yasuno(2021{\natexlab{c}})]{YasunoJSAI2021}
J.~Fujii R.~Yoshida T.~Yasuno, H.~Sugawara.
\newblock Snowy night-to-day translator and semantic segmentation label
  similarity for snow hazard indicator.
\newblock \emph{The Annual Conference of The Japanese Society for Artificial
  Intelligence(JSAI)}, 2021{\natexlab{c}}.

\bibitem[L.~Ruff and Kloft(2021)]{Ruff2021icml}
B.J. Franks K.-R.~Muller L.~Ruff, R.A.~Vandermeulen and M.~Kloft.
\newblock Rethinking assumptions in deep anomaly detection.
\newblock In \emph{The International Conference on Machine Learning (ICML),
  Workshop on Uncertainty and Robustness in Deep Learning}, 2021.

\bibitem[M.D.~Zeiler(2013)]{Zeiler2013}
R.~Fergus M.D.~Zeiler.
\newblock Visualizing and understanding convolutional networks, 2013.

\bibitem[M.T.~Ribeiro(2016)]{Ribeiro2016}
S.~Sameer M.T.~Ribeiro.
\newblock Why should i trust you? explaining the predictions of any classifier.
\newblock \emph{Knowledge Discovery and Data Mining}, 2016.

\bibitem[B.~Zhou(2015)]{Zhou2015}
A.~Lapedriza A. Oliva A.~Torralba B.~Zhou, A.~Khosla.
\newblock Learning deep features for discriminative localization, 2015.

\bibitem[R.R.~Selvaraju(2017)]{Selvaraju2017}
A.~Das R. Vedantam D. Parikh D.~Batra R.R.~Selvaraju, M.~Cogswell.
\newblock Grad-cam: Visual explanations from deep networks via gradient-based
  localization.
\newblock In \emph{Proceedings of the IEEE International Conference on Computer
  Vision (ICCV)}, 2017.

\bibitem[S.~Dorafshan(2018)]{Dorafshan2018}
M.~Maguire S.~Dorafshan, R.J.~Thomas.
\newblock Sdnet2018: An annotated image dataset for noncontact concrete crack.
\newblock \emph{Data in Brief}, 21:\penalty0 1664--1668, 2018.

\bibitem[Q.D.~Cao(2018)]{texas2018}
Y.~Choe Q.D.~Cao.
\newblock Building damage on post-hurricane satellite imagert based on
  convolutional neural networks, 2018.

\bibitem[K.~Sakurada(2015)]{sakurada2015}
T.~Okatani K.~Sakurada.
\newblock Change detection from a street image pair using cnn features and
  superpixel segmentation.
\newblock In \emph{BMVC}, 2015.

\bibitem[C.~Kyrkou(2019)]{AIDER2019}
T.~Theocharides C.~Kyrkou.
\newblock Deep-learning-based aerial image classification for emergency
  response applications using unmanned aerial vehicles.
\newblock In \emph{Workshop on Computer Vision for UAVs, CVPR}, 2019.

\end{thebibliography}

\end{document}